\documentclass[11pt,a4paper,table]{article}
\usepackage[hyperref]{emnlp2020}

\usepackage{times}
\usepackage{latexsym}

\usepackage{multirow}
\usepackage{amsmath}
\usepackage{amsfonts}
\usepackage{subfig}
\usepackage{graphicx} 
\usepackage{wasysym}
\usepackage{array}
\usepackage{float}
\usepackage{cleveref}
\usepackage{subfig}
\usepackage{booktabs,boldline}
\usepackage{arydshln}

\usepackage{colortbl}
\definecolor{gray2}{gray}{0.85}
\definecolor{gray3}{gray}{0.6}
\definecolor{gray4}{gray}{0.4}

\crefformat{section}{\S#2#1#3}
\crefformat{subsection}{\S#2#1#3}

\usepackage{mathtools}

\usepackage{microtype}

\aclfinalcopy 

\title{
Cross-Media Keyphrase Prediction: A Unified Framework with\\
Multi-Modality Multi-Head Attention and Image Wordings
}

\author{Yue Wang$^{1}$,~Jing Li$^{2}$,~Michael R. Lyu$^{1}$,~and Irwin King$^{1}$\\
$^{1}$Department of Computer Science and Engineering\\
The Chinese University of Hong Kong, HKSAR, China\\
$^{2}$Department of Computing, The Hong Kong Polytechnic University, HKSAR, China\\
$^{1}$\texttt{\{yuewang,lyu,king\}@cse.cuhk.edu.hk}\\
$^{2}$\texttt{jing-amelia.li@polyu.edu.hk}\\}

\date{}

\begin{document}
\maketitle

\begin{abstract}
Social media produces large amounts of contents every day. To help users quickly capture what they need, keyphrase prediction is receiving a growing attention. 
Nevertheless, most prior efforts focus on text modeling, largely ignoring the rich features embedded in the matching images.
In this work, we explore the joint effects of texts and images in predicting the keyphrases for a multimedia post. 
To better align social media style texts and images, we propose: (1) a novel \emph{{M}ulti-{M}odality Multi-Head Attention} (M$^3$H-Att) to capture the intricate cross-media interactions; (2) \emph{image wordings}, in forms of optical characters and image attributes, to bridge the two modalities.
Moreover, we design a novel \emph{unified} framework to leverage the outputs of keyphrase classification and generation and couple their advantages.
Extensive experiments on a large-scale dataset\footnote{Our code and dataset are released at \url{https://github.com/yuewang-cuhk/CMKP}.} newly collected from Twitter show that our model significantly outperforms the previous state of the art based on traditional co-attention networks.
Further analyses show that our multi-head attention is able to attend information from various aspects and boost classification or generation in diverse scenarios.  
\end{abstract}

\section{Introduction}

The prominent use of social media platforms (such as Twitter) 
exposes individuals with an abundance of fresh information in a wide variety of forms such as texts, images, videos, etc.
Meanwhile, the explosive growth of multimedia data has far outpaced individuals' capability to understand them.
This presents a concrete challenge to digest the massive amount of data, distill the salient contents therein, and provide users with a quicker access to the information they need when navigating a large multitude of noisy social media data.

\begin{table}
\begin{center}
\resizebox{0.5\textwidth}{!}{

\begin{tabular}{p{4.8cm}p{4.8cm}}
\textbf{Post (a)}: Contemplating the mysteries of life from inside my egg carton...\smiley{}  &
\textbf{Post (b)}: The \textit{$<$mention$>$} have the slight lead at halftime! \\
\textit{\#cat \#cats \#CatsOfTwitter} &
 \textit{\#NBAFinals}\\
\begin{minipage}{\textwidth}
      \includegraphics[width=4.8cm,height=4.8cm]{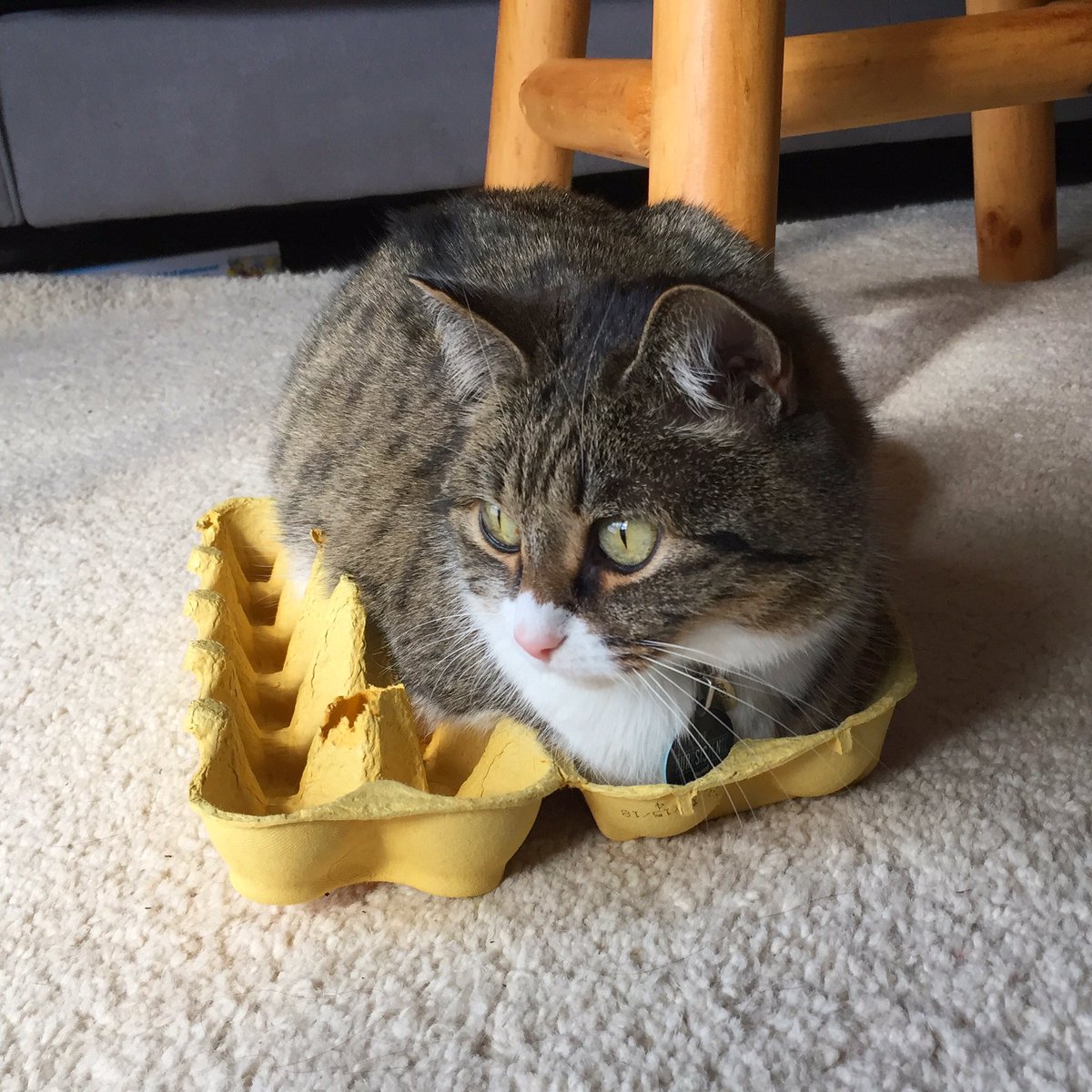}
\end{minipage}
 & 
 \begin{minipage}{\textwidth}
      \includegraphics[width=4.8cm,height=4.8cm]{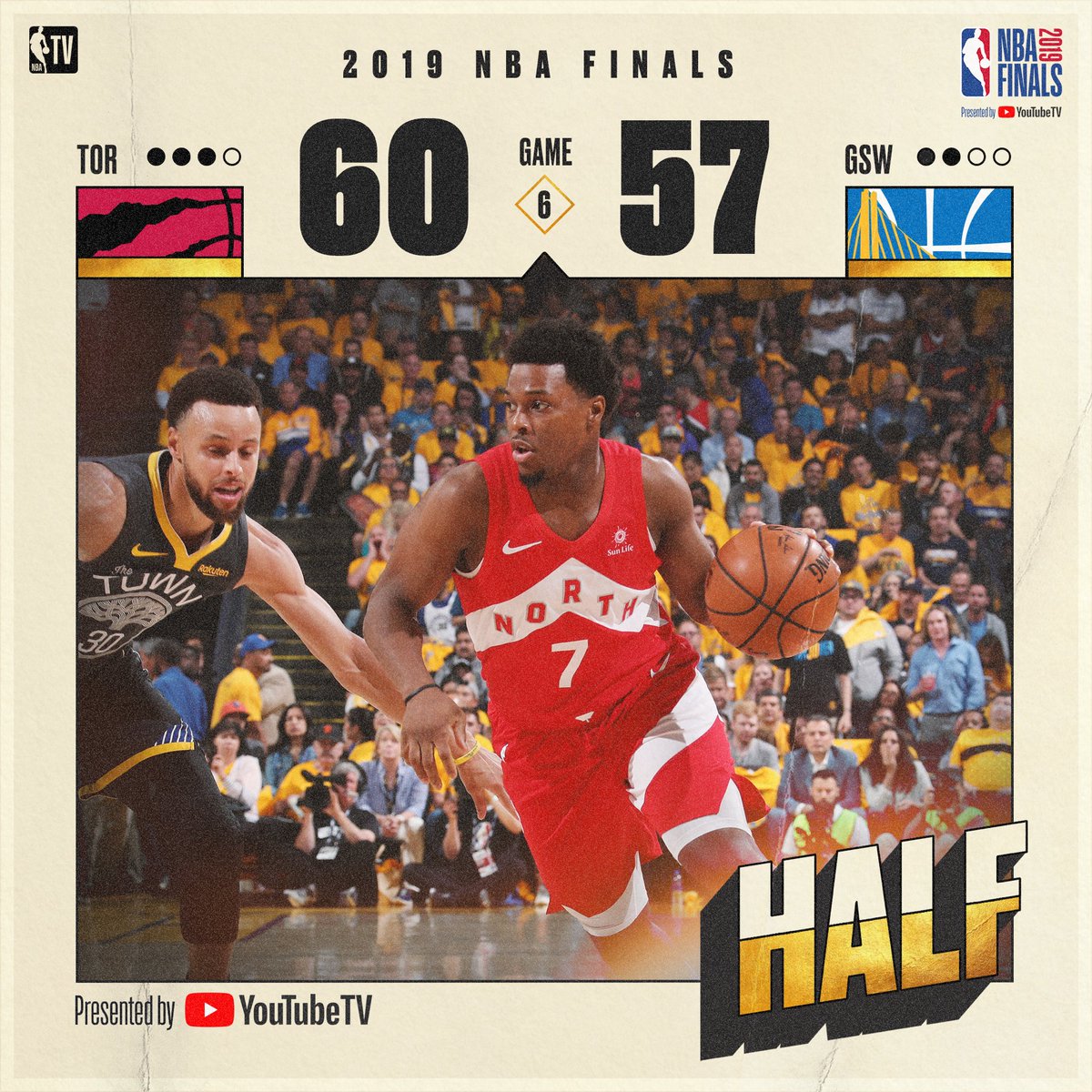}
\end{minipage} \\
\end{tabular}

}
\end{center}
\vspace{-0.5em}
\captionsetup{type=figure}
\caption{Two multimedia posts from Twitter, where texts offer limited help in identifying their keyphrases  while  images provide essential clues.
}\label{tables:intro_example}
\vspace{-0.5em}
\end{table}

To that end, extensive efforts have been made to \textbf{social media keyphrase prediction}\footnote{We consider a hashtag as a post's keyphrase annotation following the common practice~\cite{DBLP:conf/emnlp/ZhangWGH16,DBLP:conf/naacl/ZhangLSZ18}. } --- aiming to produce a sequence of words that reflect a post's key concern.
Nevertheless, previous work mostly focuses on the use of textual signals~\cite{DBLP:conf/naacl/ZhangLSZ18,wang-etal-2019-topic-aware,wang-etal-2019-microblog}, which sometimes provide limited features as social media language is essentially informal and fragmented.
To enrich the contexts, here we resort to exploiting the matching images, which are widely used in social media posts to deliver auxiliary information from authors (e.g., opinions, feelings, topics, etc.), primarily due to the flourish of mobile Internet.

To  illustrate our motivation,
Figure~\ref{tables:intro_example} shows the texts and images of two Twitter posts (tweets). 
The left is tagged with a keyphrase ``cat'', which can be clearly signaled with its image while the paired text is an anthropomorphic description and hardly unveils its real semantics.
For the right, the image depicts a basketball game scene with optical characters ``2019 NBA FINALS'', directly indicating its keyphrase, which is difficult to identify from the texts. In both examples, images play a more vital role than texts in reflecting the key information.
These points motivate our cross-media keyphrase prediction study that examines how the salient contents can be indicated by the coupled effects of post texts and their matching images.

Previous work~\cite{DBLP:conf/ijcai/ZhangWHHG17,DBLP:conf/aaai/Zhang00TY019} employs co-attention networks~\cite{DBLP:conf/nips/LuYBP16,DBLP:conf/eccv/XuS16} to encode multimedia posts, where a single attention function is concurrently performed to infer either visual or textual distributions.
We argue that they might be suboptimal to model intricate text-image associations, as a recent finding~\cite{DBLP:conf/acl/VempalaP19} points out there can be four diverse semantic relations held by images and texts on Twitter.
To allow for better modeling, in this work, 
we take advantage of the recent advances in multi-head attention~\cite{DBLP:conf/nips/VaswaniSPUJGKP17}, which is capable of learning representation from different subspaces.
We extend it to capture diverse cross-media interactions, 
named as \emph{Multi-Modality Multi-Head Attention} (M$^3$H-Att).
Moreover, to well align the images' semantics to texts', we adopt \emph{image wordings} and define two forms for that 
--- explicit \textit{optical characters}  (such as ``NBA Finals'' in post (b)) detected from the optical character reader (OCR) and implicit \textit{image attributes}~\cite{wu-etal-2006-boosting}, high-level text labels predicted to summarize the image's semantic concepts (such as a ``cat'' label for post (a)).

Furthermore, unlike prior work employing either classification ~\cite{DBLP:conf/ijcai/GongZ16} or generation models~\cite{wang-etal-2019-topic-aware},  we propose a \emph{unified} framework to couple the advantages of keyphrase classification and generation. 
Specifically, in addition to the joint training of both modules,
we further extend the copy mechanism~\cite{DBLP:conf/acl/SeeLM17} to adaptive aggregate classification outputs together with  source input tokens.
Empirical results  show that our proposed unified model  not only  keeps classification's superiority to predict common keyphrases (Figure~\ref{fig:four_sub}~(c)) while enabling keyphrase creation beyond a predefined candidate list, but also largely benefits the keyphrase prediction especially for  absent keyphrases (Figure~\ref{fig:four_sub} (b)).

For experiments, we collect a large-scale tweet dataset with texts and images, which is presented as part of our work. 
Extensive results show that our model significantly outperforms the state-of-the-art (SOTA) methods using traditional attention mechanisms.
For example, we obtain $47.06\%$ F1@1 compared with $43.17\%$ by \citet{wang-etal-2019-topic-aware} (keyphrase generation from texts only) and $42.12\%$ by~\citet{DBLP:conf/ijcai/ZhangWHHG17} (multi-modal keyphrase classification). 
We then examine how we perform to handle absent and present keyphrases, and varying keyphrase frequency and post length. 
The results indicate the consistent performance boost brought by our M$^3$H-Att design and unified framework in diverse scenarios~(\cref{ssec:freq-postlen}).
We further quantify the effects of different settings of multi-head attention and image wordings 
to see when and how they work the best~(\cref{ssec:m3h_iw}).
Lastly, we provide qualitative analysis
to interpret why our model results in superior multimedia understanding~(\cref{ssec:case_study}).





\section{Related Work}
\paragraph{Social Media Keyphrase Prediction.}
Traditional keyphrase prediction studies focus on using two-step pipeline methods: 
candidates are first extracted with handcrafted features (e.g. part-of-speech tags~\cite{DBLP:conf/dl/WittenPFGN99}) and then ranked by unsupervised~\cite{DBLP:conf/aaai/WanX08} or supervised algorithms~\cite{DBLP:conf/emnlp/MedelyanFW09}.
These methods undergo labor-intensive feature engineering and hence lead to the growing popularity of adopting data-driven neural networks. 
Specifically for social media keyphrase prediction,
most efforts are based on sequence tagging style extraction~\cite{DBLP:conf/emnlp/ZhangWGH16,DBLP:conf/naacl/ZhangLSZ18} or classification from  a predefined candidate list~\cite{DBLP:conf/ijcai/GongZ16,DBLP:conf/ijcai/ZhangWHHG17}, which cannot  produce keyphrases absent in the post or  the fixed list.

Inspired by the recent success of keyphrase generation~\cite{DBLP:conf/acl/MengZHHBC17,conf/acl/chan19keyphraseRL} for scientific articles, 
\citet{wang-etal-2019-topic-aware,wang-etal-2019-microblog} employ sequence-to-sequence (seq2seq) models to allow unseen keyphrases to be flexibly created for social media posts.
Unlike them, we propose a novel unified framework to combine the benefits of keyphrase classification and generation. 
Similar to this, \citet{DBLP:conf/naacl/ChenCLBK19} also exploit the power of  classification for keyphrase generation but in a separated retrieval manner, where we elegantly integrate them with a tailored copy mechanism and allow for the end-to-end joint training.
While most of prior work focuses on the modeling of texts, we additionally exploit the matching  images and study their coupled effects for indicating keyphrases.

\paragraph{Cross-media Research.}
We are also related to cross-media research, where texts and images are jointly exploited for a variety of applications, such as  personalized image captioning~\cite{DBLP:journals/pami/ParkKK19},  event extraction~\cite{DBLP:conf/acl/LiZZWLJC20}, sarcasm detection~\cite{DBLP:conf/acl/CaiCW19}, and text-image relation classification~\cite{DBLP:conf/acl/VempalaP19}.
Some of them have pointed out the usefulness of OCR texts~\cite{DBLP:conf/mm/ChenHK16} and image attributes~\cite{DBLP:conf/cvpr/WuSLDH16} to endow images with higher-level semantics beyond visual features, where we are the first to study how OCR texts and image attributes work together to indicate keyphrases.
Closest to our work, \citet{DBLP:conf/ijcai/ZhangWHHG17,DBLP:conf/aaai/Zhang00TY019} study multimedia hashtag classification and employ  co-attention networks~\cite{DBLP:conf/nips/LuYBP16,DBLP:conf/eccv/XuS16} to model the text-image associations, while we extend the multi-head attention~\cite{DBLP:conf/nips/VaswaniSPUJGKP17} to better capture such diverse styles of cross-media interactions. 

While multi-head attention has been widely exploited in many vision-language (VL) tasks, such as image captioning~\cite{DBLP:conf/aaai/ZhouPZHCG20}, visual question answering~\cite{tan2019lxmert,DBLP:conf/nips/LuBPL19}, and visual dialog~\cite{DBLP:conf/emnlp/KangLZ19,DBLP:conf/emnlp/vdbert20}, its potential benefit to model flexible cross-media posts has been previously ignored.
Due to the informal style in social media, cross-media keyphrase prediction brings unique difficulties mainly in two aspects: first, its text-image relationship is rather complicated~\cite{DBLP:conf/acl/VempalaP19} while in conventional VL tasks the two modalities have  most semantics shared; second, social media images usually exhibit a more diverse distribution and a much higher probability of containing OCR tokens~(\cref{sec:data}), thereby posing a hurdle for effectively processing.

\section{Our Unified Cross-Media Keyphrase Prediction Framework}

\begin{figure}
\centering
\includegraphics[scale=0.48]{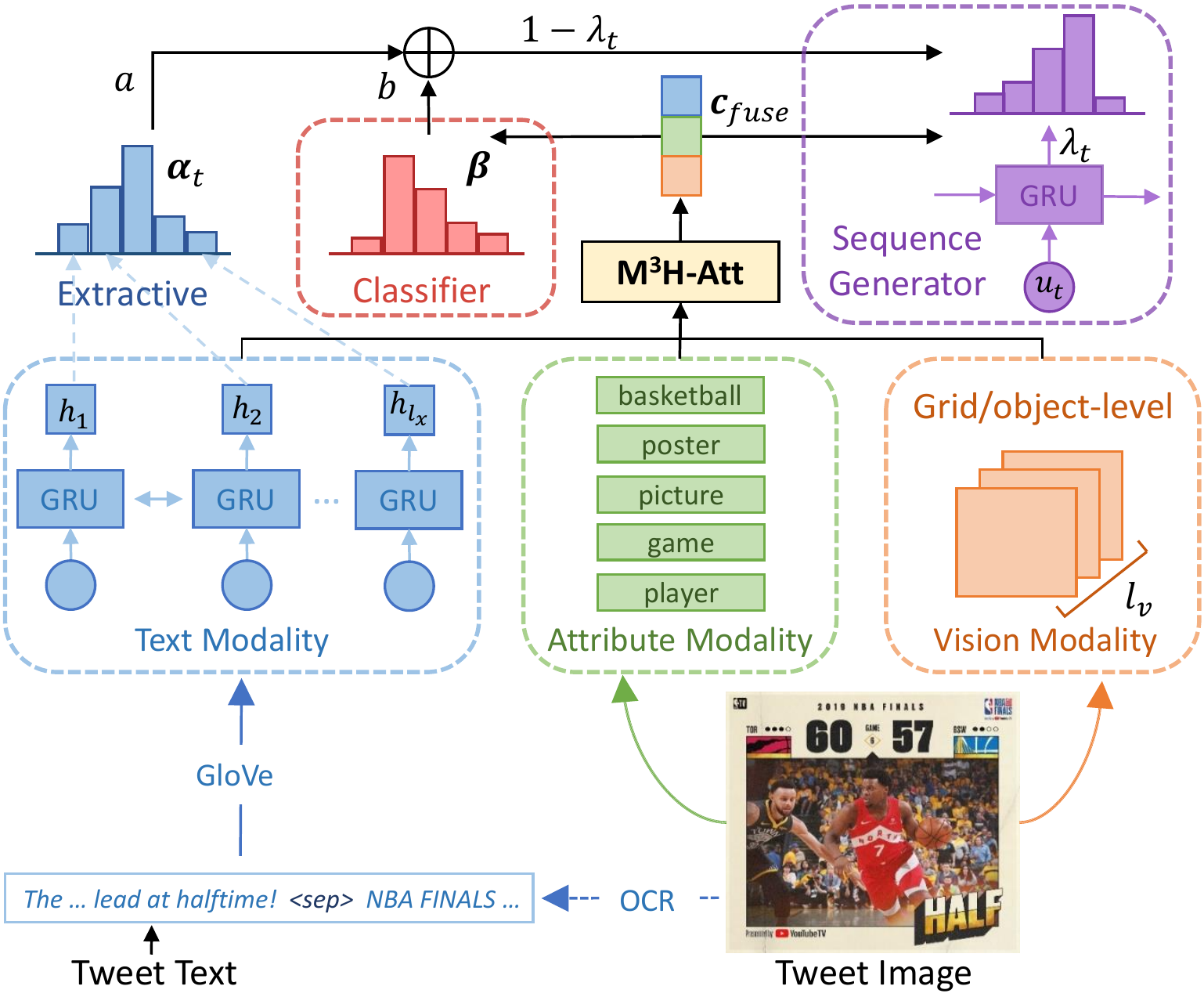}
\caption{
The overview of our unified cross-media keyphrase prediction model.}\label{fig:framework}
\vspace{-0.5em}
\end{figure}


Given a collection $\mathcal{C} $ with $|C|$ text-image post pairs $\{(\mathbf{x}^n, I^n)\}_{n=1}^{|C|}$ as input, we aim to predict a  keyphrase set 
$\mathcal{Y} =\{{\bf y}^i\}_{i=1}^{|\mathcal{Y}|}$ for each of them.
Following~\citet{DBLP:conf/acl/MengZHHBC17}, we copy the source input pair multiple times to allow each paired to have one keyphrase.
We represent each input as a triplet $(\mathbf{x}, I, \mathbf{y})$, where  $\mathbf{x}$ and  $\mathbf{y}$ are formulated as word sequences $\mathbf{x}=\langle x_1,..., x_{l_x}\rangle$ and $\mathbf{y}=\langle y_1,..., y_{l_y}\rangle$ ($l_x$ and $l_y$   denote the  number of words).

We show the  overview of our proposed cross-media keyphrase prediction model in Figure~\ref{fig:framework}.
We first encode a text-image tweet into three modalities: \textit{text}, \textit{attribute}, and \textit{vision}  (\cref{ssec:encoder}), and propose a Multi-Modality Multi-Head 
Attention (M$^3$H-Att) to capture their intricate interactions 
(\cref{ssec:model-m3h}). 
Then, we feed the learned multi-modality representations for either keyphrase classification or generation, followed with  a tailored aggregator to combine their outputs (\cref{ssec:decoder}). Lastly, the entire framework can be jointly trained via multi-task learning (\cref{ssec:training}). 

\subsection{Multi-modality Encoder}\label{ssec:encoder}

\paragraph{Learning Text Representation.}
We first embed each token $x_i$ from the input sequence into a high-dimensional vector via a pre-trained lookup table, and then employ bidirectional gated recurrent unit (Bi-GRU)~\cite{DBLP:conf/emnlp/ChoMGBBSB14} to encode the embedded input token $e(x_i)$:
\begin{align}
    \overrightarrow{\mathbf{h}_i}&=GRU(e(x_i),~~\overrightarrow{\mathbf{h}_{i-1}}),\\ \overleftarrow{\mathbf{h}_i}&=GRU(e(x_i)
    ,~~\overleftarrow{\mathbf{h}_{i+1}}).
\end{align}
Forward hidden state $\overrightarrow{\mathbf{h}_i}$ and backward one $\overleftarrow{\mathbf{h}_i}$ are later concatenated into $\mathbf{h}_i=[\overrightarrow{\mathbf{h}_i};\overleftarrow{\mathbf{h}_{i}}]$.
We employ it as the context-aware representation of $x_i$ and pack all of them
in the input sequence into a textual memory bank $\mathbf{M}_{text} =\{\mathbf{h}_i,...,\mathbf{h}_{l_x}\} \in \mathbb{R}^{l_x \times d}$, where $d$ denotes the hidden state dimension.

\paragraph{Encoding OCR Text.}
To detect optical characters from images, we use an open-source toolkit~\cite{DBLP:conf/icdar/Smith07} to extract OCR texts in  form of a word sequence.
It is then appended into the post text  with a delimited token \textit{$\langle$sep$\rangle$} to notify the change of text genres, which is shown to be a simple yet effective design to combine OCR features.

\paragraph{Learning Image Representation.}
We consider two types of  image representations: \textit{grid-level} or \textit{object-level} visual features.
For the former, we apply a pre-trained VGG-16 Net~\cite{DBLP:journals/corr/SimonyanZ14a} to extract $7 \times 7$ convolutional feature maps for each image $I$. 
For the latter, inspired by bottom-up attention~\cite{DBLP:conf/cvpr/00010BT0GZ18}, we use the Faster-RCNN~\cite{DBLP:conf/nips/RenHGS15} pre-trained on Visual Genome~\cite{DBLP:journals/ijcv/KrishnaZGJHKCKL17} to detect the objects and extract their features.
Each feature map is further transformed into a new vector $\mathbf{v}_i$ through a linear projection layer.
As such, we construct a visual memory bank  as $\mathbf{M}_{vis}=\{\mathbf{v}_1,...,\mathbf{v}_{l_v}\}\in \mathbb{R}^{l_v \times d}$, where $l_v$ denotes the number of image regions or objects.

\paragraph{Encoding Image Attribute.}
Following~\citet{DBLP:conf/acl/CaiCW19}, we first train an attribute predictor based on the Resnet-152~\cite{DBLP:conf/eccv/HeZRS16} features on MS-COCO 2014 caption dataset~\cite{DBLP:conf/eccv/LinMBHPRDZ14}.
Specifically, we extract noun and adjective tokens from the image captions as the attribute labels.
Afterward, the top five predicted attributes of each image are transformed with another linear layer to an attribute memory bank  $\mathbf{M}_{attr}=\{\mathbf{a}_1,...,\mathbf{a}_5\}\in \mathbb{R}^{5 \times d}$, which aims to  capture images' high-level semantic concepts.

\subsection{Multi-modality Multi-Head Attention}\label{ssec:model-m3h}

Our design of multi-head attention is inspired by its prototype in Transformer~\cite{DBLP:conf/nips/VaswaniSPUJGKP17}.
We extend it to capture multiple forms of cross-modality interactions for a multimedia post, which is therefore named as M$^3$H-Att, short for Multi-Modality Multi-Head Attention.
Compared to its original use as a self-attention over texts only, we instead operate on three modalities (text, attribute, and vision) in a \textit{pairwise} co-attention manner.

For each  co-attention, we perform  scaled dot attention  $\mathcal{A}$ on a set of \{\textit{Query}, \textit{Key}, \textit{Value}\}:
\begin{align}
    \mathcal{A}(Q, K, V)&=\texttt{softmax}(\frac{QK^T}{\sqrt{d_K}})V,\\
    \mathcal{A}^M(Q,K,V)&=[head_1; ...; head_H]W^O,\\
 \text{where}~~~ head_h&=\mathcal{A}(QW^Q_h, KW^K_h, VW^V_h),
\end{align}
where $W^Q_h, W^K_h, W^V_h\in\mathbb{R}^{d\times d_H}$ are learnable weights to project the query, key, and value from dimension $d$ to a lower space of $d_H$-dimension and  $H$ is the head number. 
Outputs from all the heads are  concatenated (in $\mathcal{A}^M$)  and  passed to a feedforward network  with  residual connections~\cite{DBLP:conf/eccv/HeZRS16} and layer normalization~\cite{DBLP:journals/corr/BaKH16}. 

Specifically, we employ the text features as a query to attend to the vision/attribute modality and vice versa.\footnote{We also try other combinations, e.g., M$^3$H-Att between the vision and attribute, but the improvements are negligible.} 
Here max/average-pooling is adopted to obtain one holistic query vector for each modality instead of token-level queries considering the noisy nature of social media data.
Moreover, we stack  multiple co-attention layers to  empower its modeling capability, where $L_{text}, L_{attr}, L_{vis}$ denote the number of stacked layers for text, attribute, and vision queries, respectively.
After that, the outputs from all co-attention layers are summed up with a linear multi-modal fusion layer to produce a 
context vector $\mathbf{c}_{fuse} \in \mathbb{R}^d$.
It will be  fed into a keyphrase classifier and generator for the unified prediction.
Notably, this  indicates that our  M$^3$H-Att's great potential to  serve as a generic module  for  benefiting other cross-media applications.

\begin{figure}[t]
\centering
\includegraphics[width=0.49\textwidth]{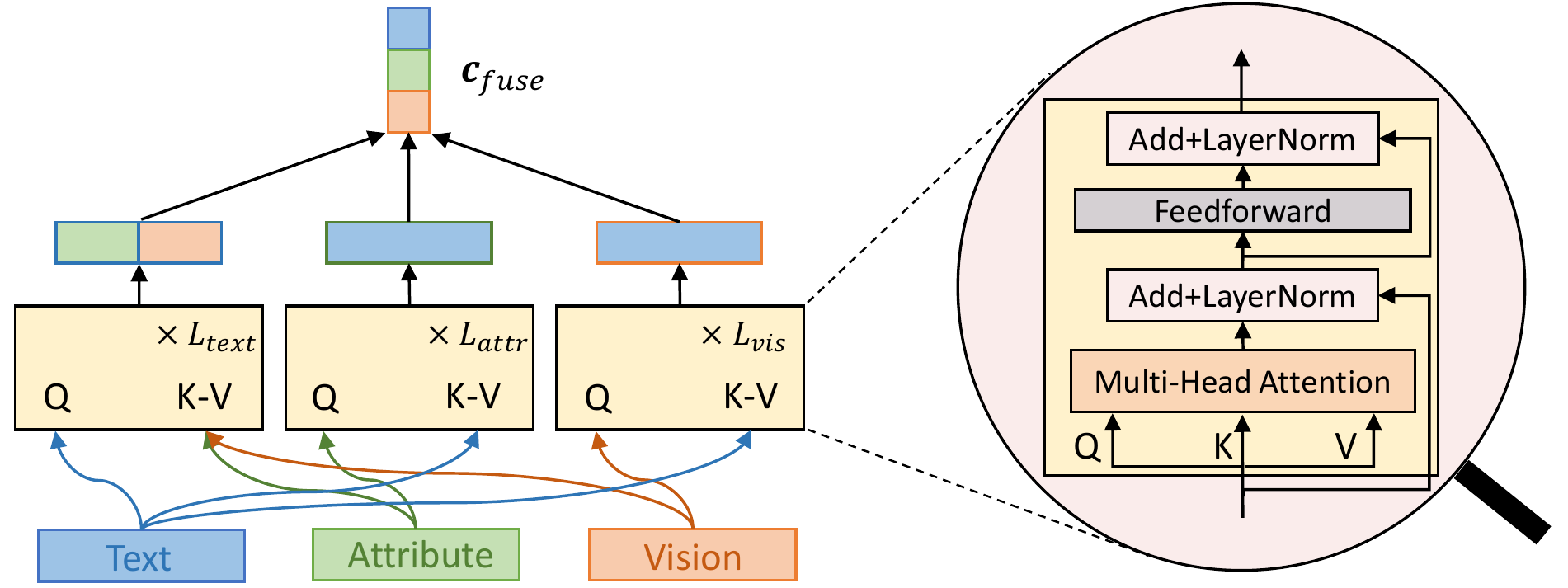}
\caption{Overview of M$^3$H-Att to fuse  multi-modal features from text, attribute, and vision modalities.}
\vspace{-0.5em}
\label{fig:m3h}
\end{figure}

\subsection{Unified Keyphrase Prediction}\label{ssec:decoder}
We describe how we  combine the keyphrase classification and generation into a unified prediction for coupling their advantages below.

\paragraph{Keyphrase Classification.}
As each keyphrase $\mathbf{y}$ usually consists of only several tokens, it can be considered as a discrete integral label and predicted it with a keyphrase classifier. 
Here we directly pass the multi-modal context vector $\mathbf{c}_{fuse}$ into a two-layer of multi-layer perceptron (MLP) and map it to the distribution over the label vocabulary $V_{cls}$:
\begin{align}
    P_{cls}(\mathbf{y}) = \texttt{softmax}(\texttt{MLP}_{cls}(\mathbf{c}_{fuse})).
\end{align}

\paragraph{Keyphrase Generation with Pointer.}
For keyphrase generation, we base on a sequence-to-sequence  framework to predict the keyphrase word sequence $\mathbf{y}=\langle y_1,..., y_{l_y}\rangle$, where the generation probability is defined as 
$\prod_{t=1}^{l_y}P(y_t\,|\,\mathbf{y}_{<t})$.

Concretely, we use an unidirectional GRU decoder to model the generation process, which emits the hidden state $\mathbf{s}_t = GRU( \mathbf{s}_{t-1}, \mathbf{u}_t) \in \mathbb{R}^d$ based on the previous hidden state $\mathbf{s}_{t-1}$ and the embedded decoder input $\mathbf{u}_t$.
The decoder state is initialized by the last hidden state $\mathbf{h}_{l_x}$ of the text encoder. 
Here  attention mechanism~\cite{DBLP:journals/corr/BahdanauCB14} is adopted to obtain a textual context $\mathbf{c}_{text}$:
\begin{align}
\mathbf{c}_{text}&=\sum_{i=1}^{l_x}\alpha_{t,i}\mathbf{h}_i,\\
\alpha_{t,i}&=\texttt{softmax} (S(\mathbf{s}_t,\mathbf{h}_i)),\label{eq:alpha} \\
S(\mathbf{s}_t,\mathbf{h}_i) &= \mathbf{v}^T_{ \alpha}~ \tanh(\mathbf{W}_{\alpha}[\mathbf{s}_t;\mathbf{h}_i] + \mathbf{b}_{\alpha}),
\end{align}
where $S(\mathbf{s}_t,\mathbf{h}_i)$ is a score function to measure the compatibility between the $t$-th word to be decoded and the $i$-th word from the text encoder.  $\mathbf{W}_{\alpha}\in \mathbb{R}^{d \times 2d},\mathbf{b}_{\alpha}, \mathbf{v}\in \mathbb{R}^{d}$ are trainable weights.

Next, we incorporate the  static multi-modal vector  $\mathbf{c}_{fuse}$ (produced by M$^3$H-Att and independent of the decoding step $t$)
to construct a context-rich representation $\mathbf{c}_t = [\mathbf{u}_t;\mathbf{s}_t;\mathbf{c}_{text} + \mathbf{c}_{fuse}]$. Based on it, we apply another MLP with softmax to produce a word distribution over  vocabulary $V_{gen}$:
\begin{equation} \label{eq:copy_score}
  P_{gen}(y_t)= \texttt{softmax}(\texttt{MLP}_{gen}(\mathbf{c}_t)).
\end{equation}

To further allow the decoder to explicitly extract words from the source post, we apply the copy mechanism~\cite{DBLP:conf/acl/SeeLM17} by 
calculating a soft switch $\lambda_t \in [0,1]$ with a sigmoid-activated MLP on $\mathbf{c}_t$.
It indicates whether to generate the word from the vocabulary $V_{gen}$ or copy it from the input sequence, where the extractive distribution is decided by the text attention weights $\alpha_{t,i}$ in Eq.~(\ref{eq:alpha}).

\paragraph{Classification Output Aggregation.}
We further extend the copy mechanism to aggregate the classification's outputs to benefit keyphrase generation.
First, we retrieve the top-K predictions from the classifier and convert each into the word sequence $\mathbf{w}=\langle w_1,..., w_{l_w}\rangle$, where $l_{w}$ is the sequence length of the combined predictions.
Then, we normalize their classification logits using softmax into a word-level distribution $\mathbf{\beta} \in \mathbb{R}^{l_{w}}$, which represents the extractive probability from the classification output.
Finally, we obtain the unified prediction via:
\begin{align}\label{eq:aggregate}
    P_{unf}&(y_t) = \lambda_t\cdot P_{gen}(y_t)~~+  \\ &(1-\lambda_t)\cdot(a\cdot\sum_{i:x_i=y_t}^{l_x}\alpha_{t,i} + b\cdot\sum_{j:w_j=y_t}^{l_{w}}\beta_j), \nonumber
\end{align}
where $a,b~(a+b=1)$ are  hyper-parameters to decide whether to copy from the input sequence or the classification outputs. 
To stabilize the aggregation of classification outputs, we warm up the classifier for several epochs first by setting $a$ to $1$ and $b$ to $0$ and then both to $0.5$ for further training.

\subsection{Joint Training Objective}\label{ssec:training}
We employ the standard negative log-likelihood loss and define the entire framework's training objective with the linear combination of the label classification loss and the token-level sequence generation loss for multitask learning:
\begin{equation}\label{eq:overall_objective}
    \mathcal{L}(\theta) = -\sum_{n=1}^N [\underbrace{\log P_{cls}(\mathbf{y}^n)}_{\text{Classification}} + \gamma \cdot \sum_{t=1}^{l_y^n}\underbrace{\log P_{unf}(y^n_t)}_{\text{Unified}}],
\end{equation}
where $N$ is size of the training text-image  pairs and $\gamma$ is a hyper-parameter to balance the two losses (empirically set to $1$) and $\theta$ denotes the   trainable parameters shared for the whole framework. Intuitively, jointly training  keyphrase classification would benefit the unified prediction by not only implicitly better parameter learning, but also explicitly providing more precise outputs to be copied to the keyphrase generator by the aggregation module.

\section{Multi-modal Tweet Dataset}\label{sec:data}
\paragraph{Data Collection and Statistics.} 
Since there are no publicly available datasets for multi-modal keyphrase annotation, we contribute a new dataset with social media posts from \textbf{Twitter}. 
Specifically, we employ the Twitter advanced search API\footnote{\url{https://twitter.com/search-advanced}} to query English tweets that contain both images and hashtags from January to June 2019. 
For keyphrases, we consider to use user-generated hashtags  following common practice~\cite{DBLP:conf/emnlp/ZhangWGH16,DBLP:conf/naacl/ZhangLSZ18}.
We further clean the raw data in the following ways:
1) we only retain tweets with one color image in JPG form; 
2) we remove tweets with less than $4$ tokens or more than $5$ hashtags to filter out noise data;
3) rare hashtags (occurring less than $10$ times) and their corresponding tweets are removed to alleviate sparsity issue;
4) we remove the duplicate tweets (e.g., retweets) and images and obtain 53,701 tweets  with each containing a distinct tweet text-image pair.
We randomly split the data into $80\%$, $10\%$, $10\%$ corresponding to training, validation, and test set. 
The data split statistics of tweet texts are displayed in Table~\ref{tables:dataset_stat}.

\begin{table}[t]\small
\begin{center}
\resizebox{0.50\textwidth}{!}{
\begin{tabular}{lccccccc}
\toprule
\multirow{2}{*}{\textbf{Split}}&\multirow{2}{*}{\#Post}  & Post & \#KP 
& \multirow{2}{*}{$|$KP$|$} &  KP  & \% of & \multirow{2}{*}{Vocab}\\ 
&& Len &/Post & & Len & occ. KP & \\

\midrule
Train& 42,959&27.26& 1.33&4,261&1.85&37.14&48,019\\
Val&5,370&26.81&1.34&2,544&1.85&36.01&16,892\\
Test &5,372&27.05&1.32&2,534& 1.86&37.45&17,021\\

\bottomrule
\end{tabular}
}
\end{center}
\vspace{-0.5em}
\caption{Data split statistics. KP: keyphrase;  $|$KP$|$: the size of unique keyphrase; \% of occ. KP: percentage of keyphrases occurring in the source post.\label{tables:dataset_stat}
}
\vspace{-0.5em}
\end{table}

\vspace{-0.3em}
\paragraph{Preprocessing.} We employ an open-source Twitter preprocessing tool~\cite{baziotis-pelekis-doulkeridis:2017:SemEval2} to tokenize the tweets, segment the hashtags, and apply common spelling corrections.
To  reduce the errors introduced by the automatic hashtag segmentation, we manually check them and construct a complete mapping list.
Following~\citet{wang-etal-2019-topic-aware}, we retain tokens in hashtags (without \# prefix) for those occurring in the middle of the posts due to their inseparable semantic roles.
We further remove all the non-alphabetic tokens and replace links, mentions (@username), digits into special tokens as \textit{$\langle$url$\rangle$}, \textit{$\langle$mention$\rangle$}, and \textit{$\langle$number$\rangle$} respectively.

\begin{figure}
\centering
\includegraphics[width=0.49\textwidth]{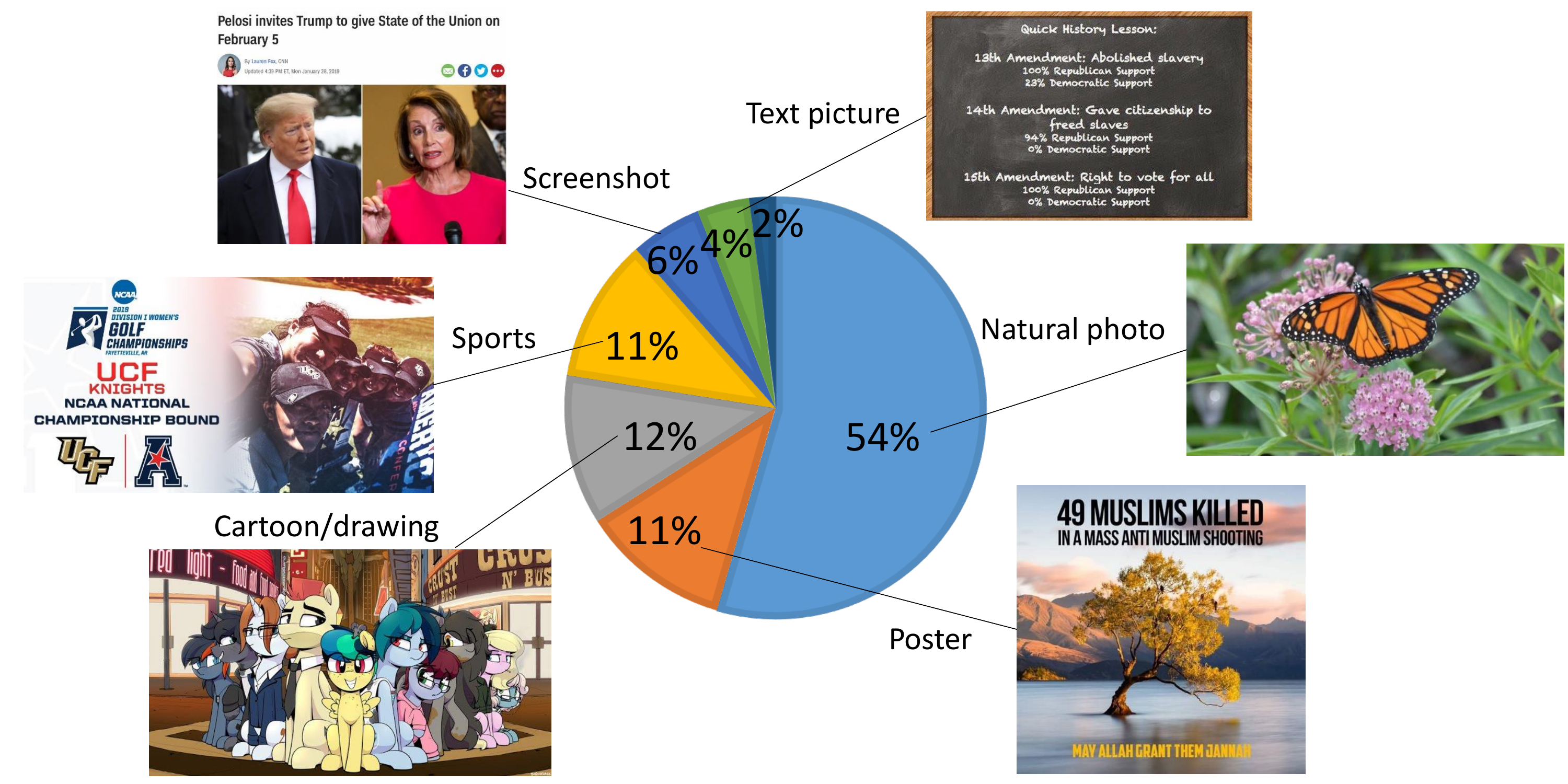}
\vspace{-0.5em}
\caption{Image type distribution of  $200$ sampled text-image tweets in our collected dataset.}\label{fig:image_types}
\vspace{-0.5em}
\end{figure}

\vspace{-0.3em}
\paragraph{Tweet Image Analysis.}
To further analyze the Twitter image characteristics, 
we sample $200$ text-image tweets and analyze their  distributions over varying types  in Figure~\ref{fig:image_types}.
We observe a diverse set of categories and only around half of the images ($54\%$) are natural photos, which is rather different from other standard image data such as MS-COCO.
Moreover, we  conduct a pilot study to categorize the text-image relations following~\citet{DBLP:conf/acl/VempalaP19} and find $52\%$ of them have either texts or images useless to represent semantics (see Figure~\ref{tables:four_association_examples} for some examples).
Such diverse category and complex text-image relationship pose unique challenges compared to traditional vision-language tasks like image captioning and visual question answering, where they focus on more natural images, and more importantly, their two modalities have most semantics shared.
To deal with this, we propose M$^3$H-Att and image wordings to better capture essential information from noisy cross-media data.

\vspace{-0.3em}
\paragraph{Image Wording Analysis.}
Here we shed light on some interesting statistics on image wordings. 
We first analyze the top $5$ attributes predicted from the images in our dataset: \textit{\{man, shirt, woman, sign, white\}}, which shows that \textit{most of the images on Twitter are about people and daily life}.
For OCR texts, we employ a widely used OCR engine Tesserocr\footnote{\url{https://pypi.org/project/tesserocr/}} to extract optical characters.
From all matching images, there are around \textbf{$35\%$} of them contain characters, significantly larger than the corresponding number in COCO images ($4\%$), indicating \textit{social media users' preference to post images containing optical characters}.  
To mitigate the effects of OCR errors,
we only consider tokens present in the vocabulary of tweet texts and
find about $17\%$ left with a median length of  $16$ tokens.
Besides, $32\%$ of the remaining data have words appearing in their corresponding keyphrases and $13\%$ contain the entire keyphrases, suggesting its potential help in keyphrase prediction.

\section{Experiments and Analyses}

\subsection{Experimental Setup}\label{ssec:setup}
\vspace{-0.3em}
\paragraph{Evaluation Metrics.}
We mainly evaluate our model with popular information retrieval metrics macro-average F1@K, where K is $1$ or $3$ as there are $1.33$ keyphrases on average per tweet (Table~\ref{tables:dataset_stat}).
To further measure the keyphrase orders (as we can generate a keyphrase ranking list with beam search), we employ mean average precision (MAP) for the top five predictions following~\citet{DBLP:conf/naacl/ChenCLBK19}.
The higher scores from all the metrics indicate better performance.
For word matchings in evaluation, we consider the results after processed with Porter Stemmer following~\citet{DBLP:conf/acl/MengZHHBC17}.

\vspace{-0.3em}
\paragraph{Comparison Models.} 
We first consider the upper-bound performance of  extractive methods, denoted as \textsc{EXT-ORACLE}. 
Then, the following baselines are compared. (1) \textbf{Image-only} models: we apply max/average pooling on the grid-level VGG features or object-level BUTD ~\cite{DBLP:conf/cvpr/00010BT0GZ18} and aggregate them for classification. 
(2) \textbf{Text-only} models: we consider  classification-based (\textsc{CLS}) or sequence generation-based (\textsc{GEN}) methods. For \textsc{CLS} models, we
consider simple max/average pooling on the text features learned from Bi-GRU encoder and the  Topic Memory Network (\textsc{TMN})~\cite{DBLP:conf/emnlp/ZengLSGLK18} (a SOTA short text classification model). 
For \textsc{GEN} models, we employ the seq2seq with attention~\cite{DBLP:journals/corr/BahdanauCB14}, copy mechanism~\cite{DBLP:conf/acl/SeeLM17}, and latent topics~\cite{wang-etal-2019-topic-aware}  (the SOTA topic-aware model for social media keyphrase generation).
(3) \textbf{Text-image} models: we consider the SOTA \textsc{CLS} model  for multi-modal hashtag recommendation~\cite{DBLP:conf/ijcai/ZhangWHHG17} using co-attention and its variant with image-attention~\cite{DBLP:conf/cvpr/YangHGDS16}, as well as  Bilinear Attention Networks (\textsc{BAN})~\cite{DBLP:conf/nips/KimJZ18} (a SOTA variant for Visual Question Answering~\cite{DBLP:conf/iccv/AntolALMBZP15}).
For our models, we first adopt the basic variants with M$^3$H-Att separately applying to either \textsc{CLS} or \textsc{GEN}. Then we additionally combine image wordings and the joint training strategy (Eq.~(\ref{eq:overall_objective})). Our full model is obtained by further aggregating the \textsc{CLS} and \textsc{GEN} outputs (Eq. (\ref{eq:aggregate})). 

\vspace{-0.3em}
\paragraph{Parameter Settings.}
We maintain a  generation  vocabulary $V_{gen}$ of $45K$ tokens and the keyphrase classification vocabulary $V_{cls}$ with 4,262 labels.
We apply $200$-d Twitter GloVe embedding~\cite{DBLP:conf/emnlp/PenningtonSM14} for encoding inputs. 
We employ two layers of Bi-GRU  for the encoder and a single layer GRU for the decoder with hidden size set to $300$.
For visual signals, we extract either $49$ grid-level VGG  $512$-d features or $36$ object-level BUTD $2048$-d features.
For the M$^3$H-Att, we employ $4$ heads with $64$-d subspace, where $4$ layers are stacked for attention to text modality, and $1$ layer for vision or attribute modality. 
In training, we set the loss coefficient $\gamma=1$ and employ Adam optimizer~\cite{DBLP:journals/corr/KingmaB14} with a learning rate  as $0.001$.
We decay it by $0.5$ if validation loss does not drop and apply gradient clipping with the max gradient norm as $5$. 
Early stop~\cite{caruana2001overfitting} is adopted via monitoring the change of validation loss. 
For inference, we employ beam search with beam size set to $10$ to generate a ranking list of keyphrases.
For the baselines, we re-implement \textsc{CLS-IMG-ATT} and \textsc{CLS-CO-ATT}, and employ the released codes to produce results for \textsc{CLS-TMN}\footnote{\url{https://github.com/zengjichuan/TMN}}, \textsc{GEN-TOPIC}\footnote{\url{https://github.com/yuewang-cuhk/TAKG}}, and \textsc{CLS-BAN}\footnote{\url{https://github.com/jnhwkim/ban-vqa}}.

\begin{table}[t]
\begin{center}
\resizebox{0.5\textwidth}{!}{
\begin{tabular}{p{0.033\textwidth}lccc}
\toprule

&\textbf{Models}
& \textbf{F1@1} & \textbf{F1@3} & \textbf{MAP@5} \\
\midrule

&\textsc{EXT-ORACLE} &39.50&23.20&39.26\\

\midrule
\multirow{4}{*}{\parbox[t]{2mm}{\rotatebox[origin=c]{90}{\textbf{Image-only}}} $\begin{dcases*} \\ \\ \\  \end{dcases*}$} 
&\textsc{CLS-VGG-MAX} 
&14.20\textsubscript{35}&12.20\textsubscript{24}&17.68\textsubscript{31}\\

&\textsc{CLS-VGG-AVG} 
&15.69\textsubscript{21}&13.67\textsubscript{06}&19.70\textsubscript{20}\\

&\textsc{CLS-BUTD-MAX} 
&17.65\textsubscript{32}&15.00\textsubscript{21}&21.77\textsubscript{29}\\ 

&\textsc{CLS-BUTD-AVG} 
&20.02\textsubscript{27}&16.97\textsubscript{06}&24.73\textsubscript{11}\\

\midrule

\multirow{6}{*}{\parbox[t]{2mm}{\rotatebox[origin=c]{90}{\textbf{Text-only}}} $\begin{dcases*} \\ \\ \\ \\ \\  \end{dcases*}$} 
&\textsc{CLS-AVG} 
&35.96\textsubscript{11} &27.59\textsubscript{05} &41.84\textsubscript{14}\\

&\textsc{CLS-MAX} 
&38.33\textsubscript{47} &28.84\textsubscript{09} &44.15\textsubscript{34}\\

&\textsc{CLS-TMN}
&40.33\textsubscript{39}& 30.07\textsubscript{28} &46.28\textsubscript{27}\\ 

\cdashline{2-5}
&\textsc{GEN-ATT}
&38.36\textsubscript{28}&27.83\textsubscript{15}&43.35\textsubscript{20}\\
&\textsc{GEN-COPY}
&42.10\textsubscript{19}&29.91\textsubscript{30}&46.94\textsubscript{35}\\
&\textsc{GEN-TOPIC}
&43.17\textsubscript{24} &30.73\textsubscript{13}&48.07\textsubscript{23}\\

\midrule
\multirow{10}{*}{\parbox[t]{2mm}{\rotatebox[origin=c]{90}{\textbf{Text-Image}}} $\begin{dcases*} \\ \\  \\ \\ \\ \\ \\ \\ \\ \end{dcases*}$} 
&\textsc{CLS-BAN}
& 38.73\textsubscript{18} &29.68\textsubscript{23} &45.03\textsubscript{15}\\
&\textsc{CLS-IMG-ATT}
&41.48\textsubscript{33} & 31.22\textsubscript{14} &47.93\textsubscript{34}\\
&\textsc{CLS-CO-ATT}
& 42.12\textsubscript{38} &31.55\textsubscript{33} &48.39\textsubscript{34}\\

\cmidrule(lr){2-5}
&\textsc{CLS-M\textsuperscript{3}H-ATT}~(ours) 
&44.11\textsubscript{17} &31.47 \textsubscript{14} &49.45\textsubscript{11}\\
&~~~+ image wording  
&44.46\textsubscript{12} & 32.82\textsubscript{24}& 50.39\textsubscript{15} \\
&~~~+ joint-train  
&45.16\textsubscript{09} &33.27\textsubscript{10} &51.48\textsubscript{11}\\

\cdashline{2-5}
&\textsc{GEN-M\textsuperscript{3}H-ATT}~(ours) 
&44.25\textsubscript{05} &31.58\textsubscript{13} &49.35\textsubscript{10}\\
&~~~+ image wording  
&44.56\textsubscript{09} & 31.77\textsubscript{23}& 49.95\textsubscript{22}\\
&~~~+ joint-train  
&45.69\textsubscript{17} &32.78\textsubscript{09} &51.37\textsubscript{12}\\
\cdashline{2-5}
&\textsc{GEN-CLS-M\textsuperscript{3}H-ATT}~(ours) 
&\textbf{47.06}\textsubscript{04} & \textbf{33.11}\textsubscript{01} &\textbf{52.07}\textsubscript{03}\\

\bottomrule
\end{tabular}
}
\end{center}
\vspace{-0.5em}
\caption{Comparison results (in \%) displayed with average scores from $5$ random seeds. Our  \textsc{GEN-CLS-M$^3$H-ATT} significantly outperforms all the comparison models (paired t-test $p<0.05$). Subscripts denote the standard deviations (e.g., 47.06\textsubscript{04} $\Rightarrow$ 47.06$\pm$0.04).
}\label{tables:main_exp_new}
\vspace{-0.5em}
\end{table}

\subsection{Main Comparison Results}\label{ssec:main-results}

We first report the main comparison results in Table~\ref{tables:main_exp_new} and draw the following observations:

$\bullet$~\textit{Textual features are more important than visual signals}. It is seen from the text-only models' better performance compared with their counterparts relying solely on images. 
For image-only models, we find that object-level BUTD outperforms grid-level VGG, while for pooling methods, average pooling works better for visual signals while max pooling is more suitable for texts.\footnote{In experiments, we find that VGG works better than BUTD features for M$^3$H-Att in our variants. We show results with the better setting without otherwise specified.} 

$\bullet$ \textit{Vision modality can provide complementary information to the text.}
Most models considering cross-media signals perform better than text-only and image-only baselines.
An exception is observed on the classification models with traditional attention, where the best F1@1 score $42.12$ from \textsc{CLS-CO-ATT} is still less than the  text-only model \textsc{GEN-TOPIC}'s  $43.17$.
This indicates the limitation of traditional co-attention to well exploit multi-modal features from social media.

$\bullet$ \textit{Both M$^3$H-Att and image wordings are helpful to encode social media features.} We find that both M$^3$H-Att and image wordings contribute to the performance boost of keyphrase classification or generation or their joint training results, which showcase their ability to handle multi-modality data from social media.
We will discuss more in \cref{ssec:m3h_iw}.

$\bullet$ \textit{Our output aggregation strategy is effective.} Seq2seq-based keyphrase generation models (especially armed with the copy mechanism to enable better extraction capability) perform better than most classification models and even upper bound results of extraction models.
It is probably because of the high absent keyphrase rate and the large size of keyphrases (Table~\ref{tables:dataset_stat}) exhibited in the noisy social media data. Nevertheless, \textsc{GEN-CLS-M$^3$H-ATT}, coupling advantages of classification and generation, obtains the best results ($47.06$ F1@1), dramatically outperforms the SOTA text-only model ($43.17$)  and text-image one ($42.12$).

\begin{figure}
\centering
\includegraphics[width=0.49\textwidth
]{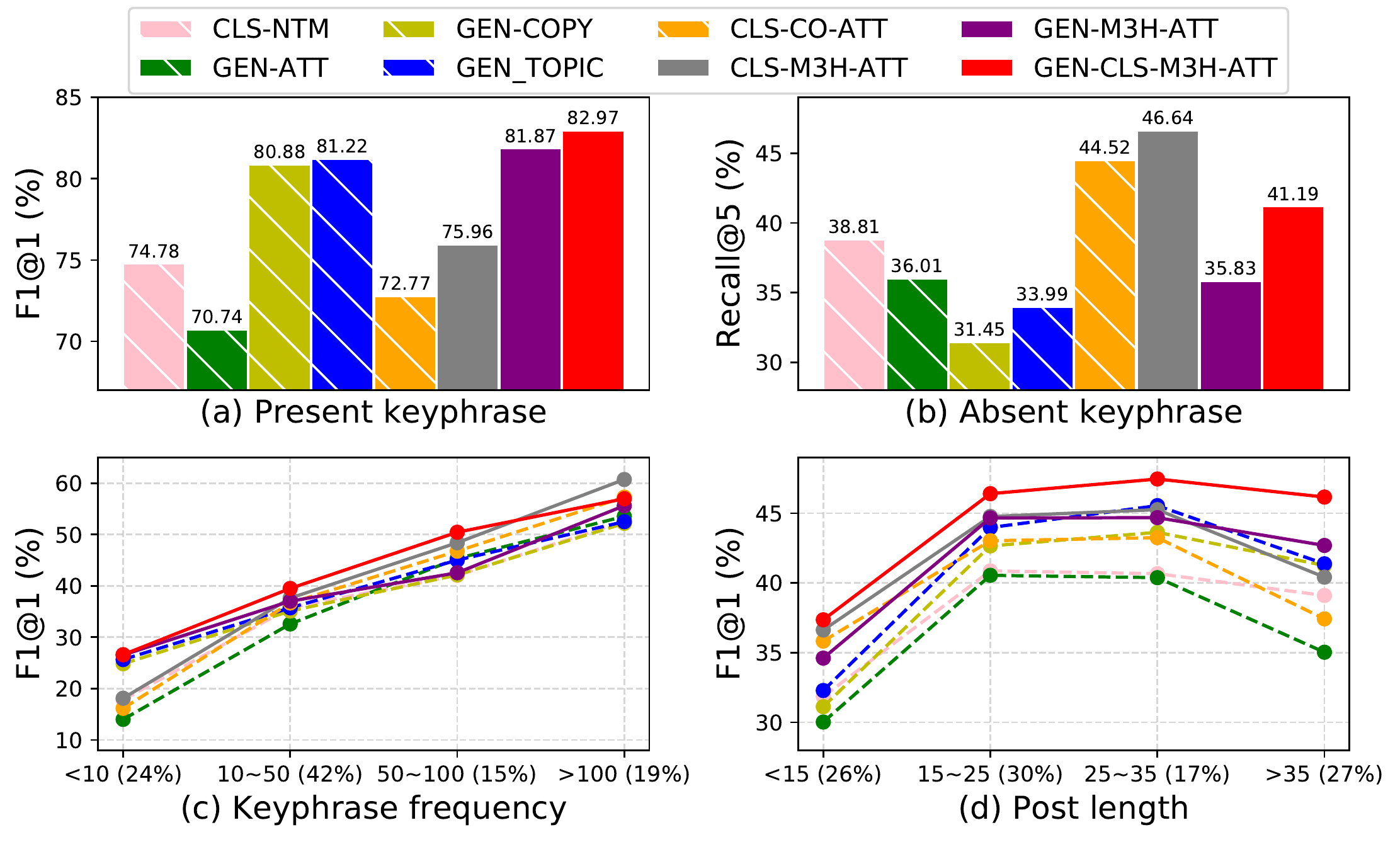}
\vspace{-1em}
\caption{
Model comparison over: (a) present keyphrases, (b) absent keyphrases, (c) varying keyphrase frequency, and (d) varying post length. 
Striped bars or dashed lines denote previous models while solid ones denote ours.
}
\vspace{-1em}
\label{fig:four_sub}
\end{figure}

\subsection{Quantitative Analyses}\label{ssec:freq-postlen}
We examine how our models  perform in diverse scenarios: present vs. absent keyphrases and varying keyphrase frequency and post length in Figure~\ref{fig:four_sub}. 
\vspace{-0.2em}
\paragraph{Present vs. Absent Keyphrases.}
We report the F1@1 for evaluating  present keyphrases and recall@5 for absent keyphrases.
As shown in  Figure~\ref{fig:four_sub} (a-b), generation models with copy mechanism consistently outperform classification models for present keyphrases, while the latter works better for absent keyphrases.
Nonetheless, our output aggregation strategy is able to cover generation models' inferiority for absent keyphrases and exhibits much better results ($41.19$ vs. $35.83$ recall@5 score) from \textsc{GEN-CLS-M$^3$H-Att} than \textsc{GEN-M$^3$H-Att}.
Besides, visual signals are helpful for  generation and classification models in predicting both present and absent keyphrases, where a larger boost is observed for the latter, probably owing to the inadequate clues available from texts.

\vspace{-0.2em}
\paragraph{Keyphrase Frequency.}
From Figure~\ref{fig:four_sub} (c), we observe better F1@1 from all models to produce more frequent keyphrases, because common keyphrases allow better representation learning from more training instances.
For extremely rare keyphrases (occur $<10$ times in training), generation models with copy mechanisms exhibit better capability to handle them than classification ones.

\vspace{-0.2em}
\paragraph{Post Length.}
From Figure~\ref{fig:four_sub} (d), we observe that longer post length does not guarantee better performance and the best results are obtained for posts with $15\sim35$ tokens.
It might be attributed to the noisy nature of social media data --- longer posts provide both richer contents and more noise.
For the posts with $<15$ tokens, all multi-modal methods perform better than the text-only ones, indicating that the image modality plays a more important when texts contain limited features.

\subsection{Analyses of M$^3$H-Att and Image Wording}\label{ssec:m3h_iw}
We proceed to quantify the effects of different settings in M$^3$H-Att and image wording. 

\begin{table}[t]
\begin{center}
\resizebox{0.5\textwidth}{!}{
\begin{tabular}{ccccccccccccc}
\toprule

\multirow{2}{*}{\textbf{\# Layer}}
&  \multicolumn{3}{c}{\textbf{2 Head}} &  \multicolumn{3}{c}{\textbf{4 Head}} &  \multicolumn{3}{c}{\textbf{8 Head}} &  \multicolumn{3}{c}{\textbf{12 Head}}\\

\cmidrule(lr){2-4}
\cmidrule(lr){5-7}
\cmidrule(lr){8-10}
\cmidrule(lr){11-13}

& 64-d & 128-d & 256-d  & 64-d & 128-d & 256-d  & 64-d & 128-d & 256-d & 64-d & 128-d & 256-d \\

\midrule
1 
& \cellcolor{gray2}42.06& \cellcolor{gray2}43.32& \cellcolor{gray2}43.01
& \cellcolor{gray2}43.11& \cellcolor{gray2}43.98&\cellcolor{gray2}43.63
& \cellcolor{gray2}43.75&\cellcolor{gray3}44.18&\cellcolor{gray2}43.43 
& \cellcolor{gray2}43.48& \cellcolor{gray2}43.81& \cellcolor{gray2}43.53\\
2 
& \cellcolor{gray2}43.22&\cellcolor{gray3}44.36&\cellcolor{gray3}44.26
&\cellcolor{gray3}44.27&\cellcolor{gray3}44.38&\cellcolor{gray3}44.27
&\cellcolor{gray3}44.58&\cellcolor{gray3}44.59& \cellcolor{gray2}43.12 
& \cellcolor{gray4}45.05&38.16&39.97\\
3 
& \cellcolor{gray2}43.51&\cellcolor{gray3}44.23& \cellcolor{gray2}43.62
&\cellcolor{gray3}44.50&\cellcolor{gray3}44.25& \cellcolor{gray2}43.00
& \cellcolor{gray3}44.70& \cellcolor{gray2}43.27&36.05
&\cellcolor{gray3}44.49&35.70&31.35 \\
4 
&\cellcolor{gray3}44.38&\cellcolor{gray3}44.42&31.72
& \cellcolor{gray4}45.29&36.03&30.47
& 37.17&32.73&31.69 
&37.85&34.99&30.91 \\
\bottomrule
\end{tabular}
}
\end{center}
\vspace{-0.8em}
\caption{Analysis of  M$^3$H-Att with various stacked layer number, head number, and subspace dimension.
}\label{tables:exp-m3h}
\vspace{-0.8em}
\end{table}

\begin{table}[t]
\begin{center}
\resizebox{0.5\textwidth}{!}{
\begin{tabular}{lccccccccccc}
\toprule

\multirow{2}{*}{\textbf{Models}}
&  \multicolumn{3}{c}{\textbf{No Image Wording}} &  \multicolumn{4}{c}{\textbf{Add OCR}} &  \multicolumn{4}{c}{\textbf{Add Attribute}} \\

\cmidrule(lr){2-4}
\cmidrule(lr){5-8}
\cmidrule(lr){9-12}

& Full & OCR  & Attr &Full & $\Delta$ (\%) & OCR & $\Delta$ (\%) & Full & $\Delta$ (\%) & Attr & $\Delta$ (\%) \\

\midrule

\textsc{CLS-MAX}
&38.31&36.11& \multicolumn{1}{c|}{32.04} 
&38.75 & +1.1 &40.67 & +12.6
& \multicolumn{1}{|c}{41.09} & +7.3 &37.87 & +18.2 \\

\textsc{GEN-COPY}
&42.01&40.81&\multicolumn{1}{c|}{35.55}
&42.86 & +2.0 &43.58 &  +6.8
&\multicolumn{1}{|c}{43.11}  & +2.6 &38.10  &+7.2\\

\textsc{CLS-M\textsuperscript{3}H-ATT}
&44.19 &42.93&\multicolumn{1}{c|}{36.93}
&44.27 &+0.2 & 46.53 &+8.4
&\multicolumn{1}{|c}{44.38}  &+0.4 & 38.73 & +4.9\\

\textsc{GEN-M\textsuperscript{3}H-ATT}
&44.33&43.26&\multicolumn{1}{c|}{35.93}
&44.48  &+0.3&  46.31 &+7.1
&\multicolumn{1}{|c}{44.77}  &+1.0 & 39.90 &+11.0\\

\bottomrule

\end{tabular}
}
\end{center}
\vspace{-0.8em}
\caption{
F1@1 over three test sets with various settings: no image wording, adding either OCR or attribute.
$\Delta$: the relative improvements over no image wording.}\label{tables:ocr_attr}
\vspace{-0.5em}
\end{table}

\vspace{-0.2em} 
\paragraph{M$^3$H-Att Analysis.}
We investigate how various configurations ($L_{vis}\in\{1,2,3,4\}$, $H\in\{2,4,8,12\}$,  $d_H\in \{64,128,256\}$ ) of our M$^3$H-Att affect the prediction results in Table~\ref{tables:exp-m3h}.
Here we only show the classification results (and similar trends are observed from generation).
We notice that more complex models do not always present better results and even render performance deteriorate in some cases due to the overfitting issue.
The best performance is attained by $4$ stacked layers of $4$ heads with  a $64$-d subspace.

\vspace{-0.2em}
\paragraph{Image Wording Analysis.} 
To examine image wording effects, we compare four models in three settings: no image wording, OCR (only), and image attributes (only) in Table~\ref{tables:ocr_attr}.
The results are shown in three test sets:  
the entire test set (Full), the $889$ subset instances with OCR tokens (OCR), and the $266$ ones containing keyphrases from ImageNet labels (Attr)~\cite{DBLP:journals/ijcv/RussakovskyDSKS15}.
For the \textsc{CLS-MAX} and \textsc{GEN-COPY}, we add attributes by using its max-pooled features to attend the text memory, which is later used for prediction.

We observe that either OCR texts or image attributes contribute to better F1@1 on the entire test set for all chosen models, while much more performance gain can be observed on their subsets with OCR texts or ImageNet keyphrases, indicating that images with optical characters and natural styles can benefit more from image wordings.\footnote{Here we assume that multimedia posts with ImageNet keyphrases have a higher probability to contain natural photos.}

\begin{figure}
\centering
\includegraphics[width=0.49\textwidth, trim={0 0.5cm 0 0}]{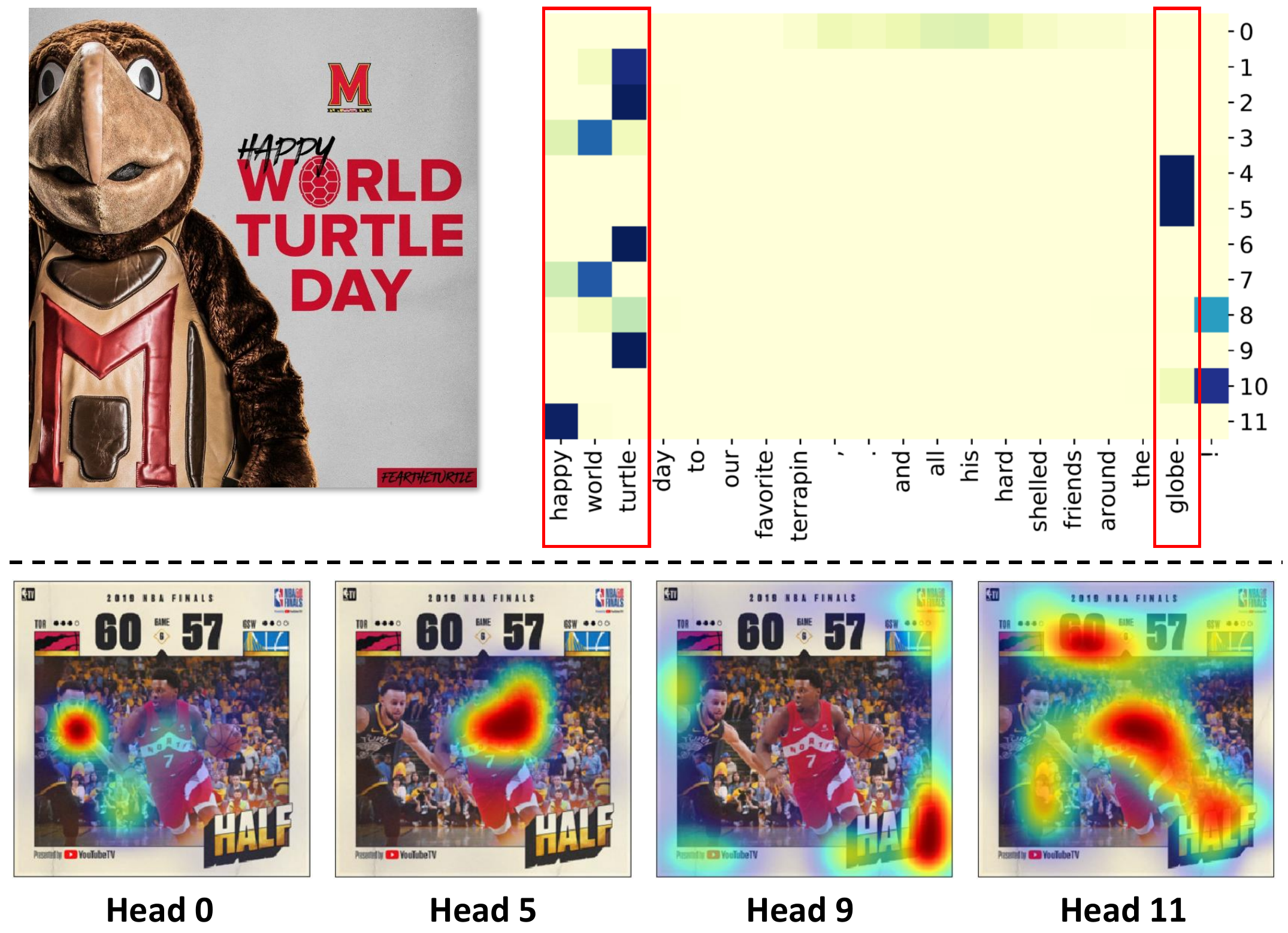}
\vspace{-0.5em}
\caption{Attention weight visualization of M$^3$H-Att for two example posts with image-to-text (top)  and text-to-image attention (bottom). Best viewed in color.}\label{fig:attn_vis}
\vspace{-0.5em}
\end{figure}

\subsection{Qualitative Analysis}\label{ssec:case_study}
To explore whether M$^3$H-Att is able to attend different aspects from the image, we probe into its attention weights via heatmap visualization in Figure~\ref{fig:attn_vis}.
Here \textsc{CLS-M\textsuperscript{3}H-ATT} is employed with a single layer of $12$ heads, whose image-to-text and text-to-image attention are examined.
The top figure shows that all its heads attend to the text based on the visual cues, where some attend to ``turtle'' while others attend to ``world'' and ``globe'' with various emphasis.
Interestingly, Head 11 highlights the ``happy'' token, which also appears in the image.
For the text-to-image attentions (bottom), we find some heads tend to highlight the specific local objects, such as the two players by Head 0 and 5 and the textual regions by Head 9, while some capture a more global view of the image like Head 11.
More examples are shown in Figure~\ref{fig:more_attn_cases}.

\begin{table}[t]
\begin{center}
\resizebox{0.5\textwidth}{!}{
\begin{tabular}{p{4.5cm}p{4.5cm}p{4.5cm}}
\textbf{Post (a)}: Contemplating the mysteries of life from inside my egg carton...\smiley{} 
 & 
\textbf{Post (b)}: Epic Texas \textit{\#sunset} from NNE Bastrop County TX. 
@TxStormChasers 
&
\textbf{Post (c)}: Your plastic bag ends up somewhere, and sometimes, it goes to the ocean.~~~\textit{\#WorldOceansDay}
\\
\begin{minipage}{\textwidth}
      \includegraphics[width=4.7cm,height=3.7cm]{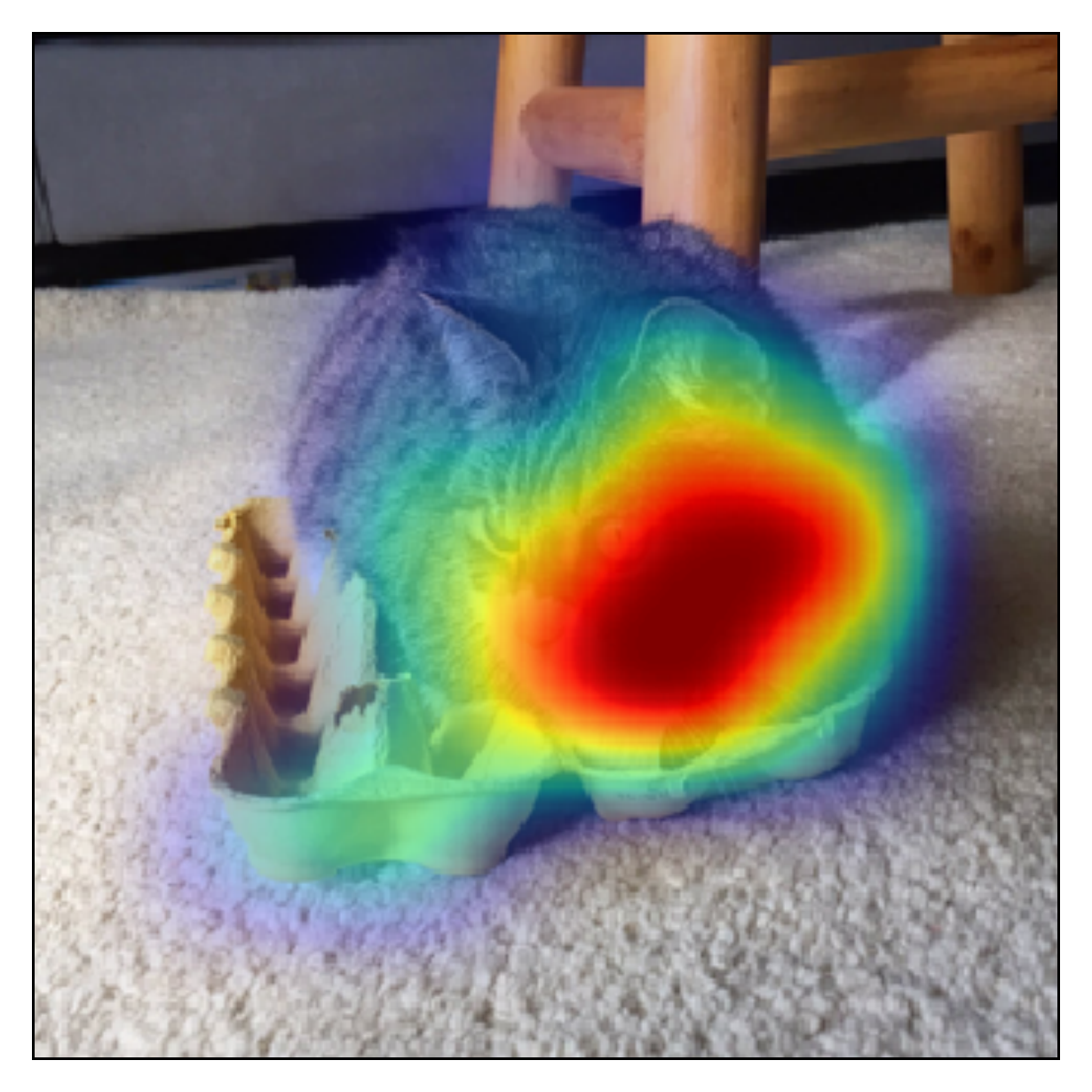}
\end{minipage}
&
 \begin{minipage}{\textwidth}
      \includegraphics[width=4.5cm,height=3.5cm]{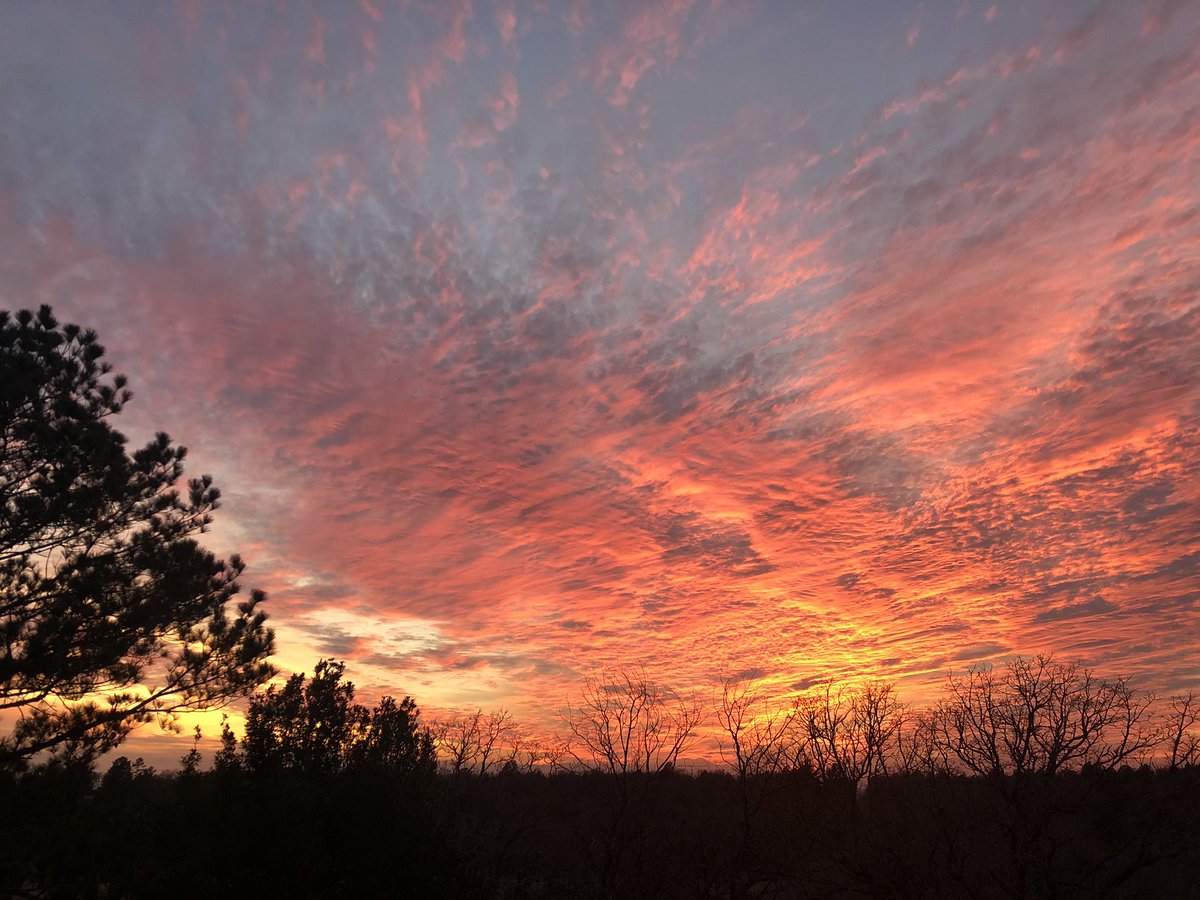}      
\end{minipage} 
&
\begin{minipage}{\textwidth}
      \includegraphics[width=4.5cm,height=3.5cm]{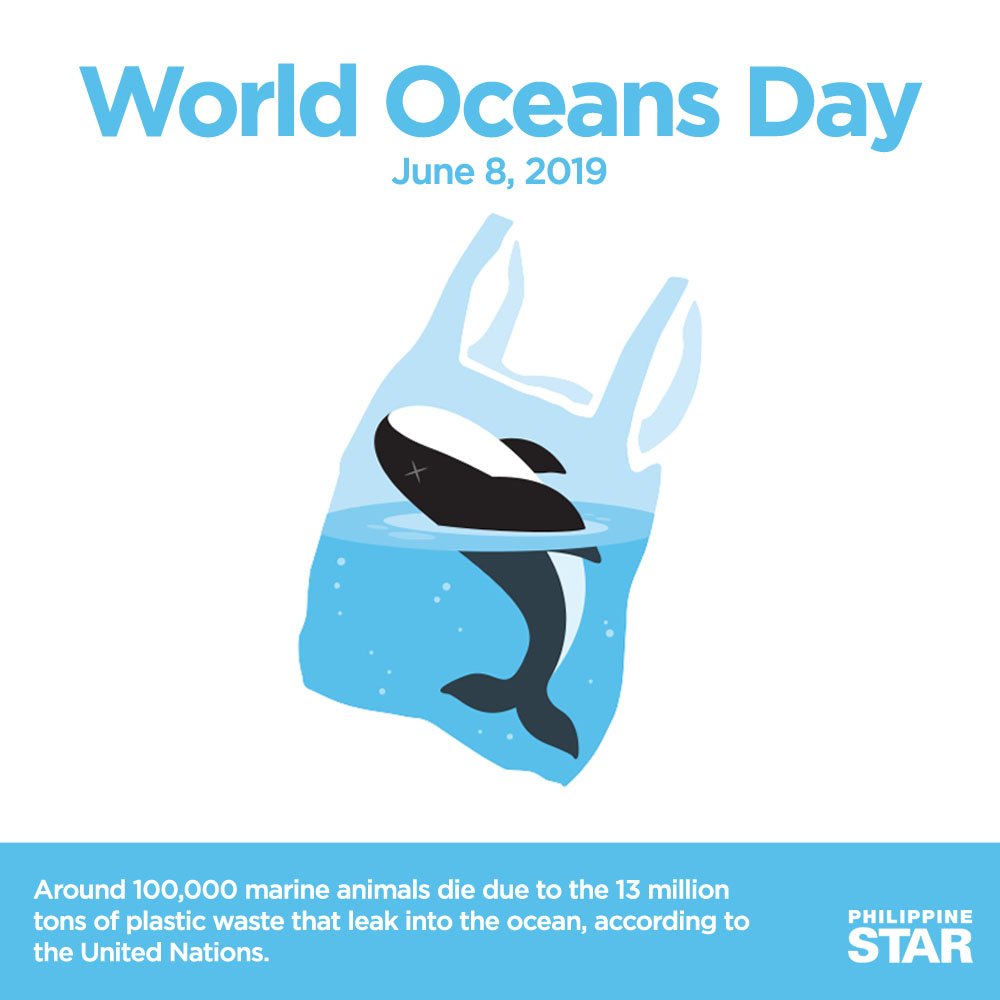}
\end{minipage}
\\
\multicolumn{1}{c}{(\textcolor{blue}{cat yellow grey bananas})}
&
\multicolumn{1}{c}{(\textcolor{blue}{sky sun sunset field})}
&
\multicolumn{1}{c}{(\textcolor{purple}{world oceans day June 8})}
\\
\textsc{gen-copy}: star wars 

\textsc{cls-co-att}: \textbf{cats of twitter}

Our: \textbf{cats of twitter}
&
\textsc{gen-copy}: storm hour

\textsc{cls-co-att}: storm hour

Our: \textbf{sunset}
&
\textsc{gen-copy}: plastic fandom

\textsc{cls-co-att}: plastic

Our: \textbf{world oceans day}
\end{tabular}
}
\end{center}
\captionsetup{type=figure}
\vspace{-0.8em}
\caption{Tweet image's effects for keyphrase prediction. \textcolor{blue}{Blue tokens} are the top $4$ attributes and \textcolor{purple}{purple ones} are OCR tokens. Correct predictions are in \textbf{bold}.}\label{fig:case_study}
\vspace{-0.5em}
\end{table}

We further illustrate how images (visual signals, image attributes, and  OCR tokens) help cross-media keyphrase prediction by analyzing their predictions in Figure~\ref{fig:case_study}. 
In post (a),  visual features help both \textsc{CLS-CO-ATT} and our model correctly predict its keyphrase, where our model precisely attends the cat's face (key region reflecting the image's semantics).
Without such context, \textsc{GEN-COPY} wrongly predicts \textit{``star wars''}, which might be caused by the misleading token \textit{``mysterious''} in the texts.
Besides, the cat keyphrase  is also revealed in the top  predicted attribute. 
In post (b-c), only our model with image wordings makes  correct predictions, where we observe that the ground-truth keyphrases directly appear in the attributes or  OCR texts. See  Figure~\ref{tables:OCR_cases} and \ref{tables:attr_cases} for more  examples.

\section{Conclusion}

This paper studies cross-media keyphrase prediction on social media and presents a unified framework to couple the advantages of generation and classification models for this task.
Moreover, we propose a novel \emph{Multi-Modality Multi-Head Attention} 
to capture the dense interactions between texts and images, where image wordings explicit in optical characters and implicit in image attributes are further exploited to bridge their semantic gap. Experimental results on a large-scale newly-collected Twitter corpus show that our model significantly outperforms SOTA  generation and classification models with  traditional attention mechanisms. 

\section*{Acknowledgements}
We thank  Hou Pong Chan, Jian Li, and our anonymous reviewers for their insightful feedback on our paper.
This work was partially supported by the Research Grants Council of the
Hong Kong Special Administrative Region, China (No. CUHK 14210717, General Research Fund; No. CUHK 2410021, Research Impact Fund, R5034-18). Jing Li is supported by the Hong Kong Polytechnic University internal funds (1-BE2W and 1-ZVRH) and NSFC Young Scientists Fund 62006203.
\bibliography{emnlp2020}
\bibliographystyle{acl_natbib}

\newpage
\appendix

\begin{figure*}
\centering
\includegraphics[width=0.99\textwidth]{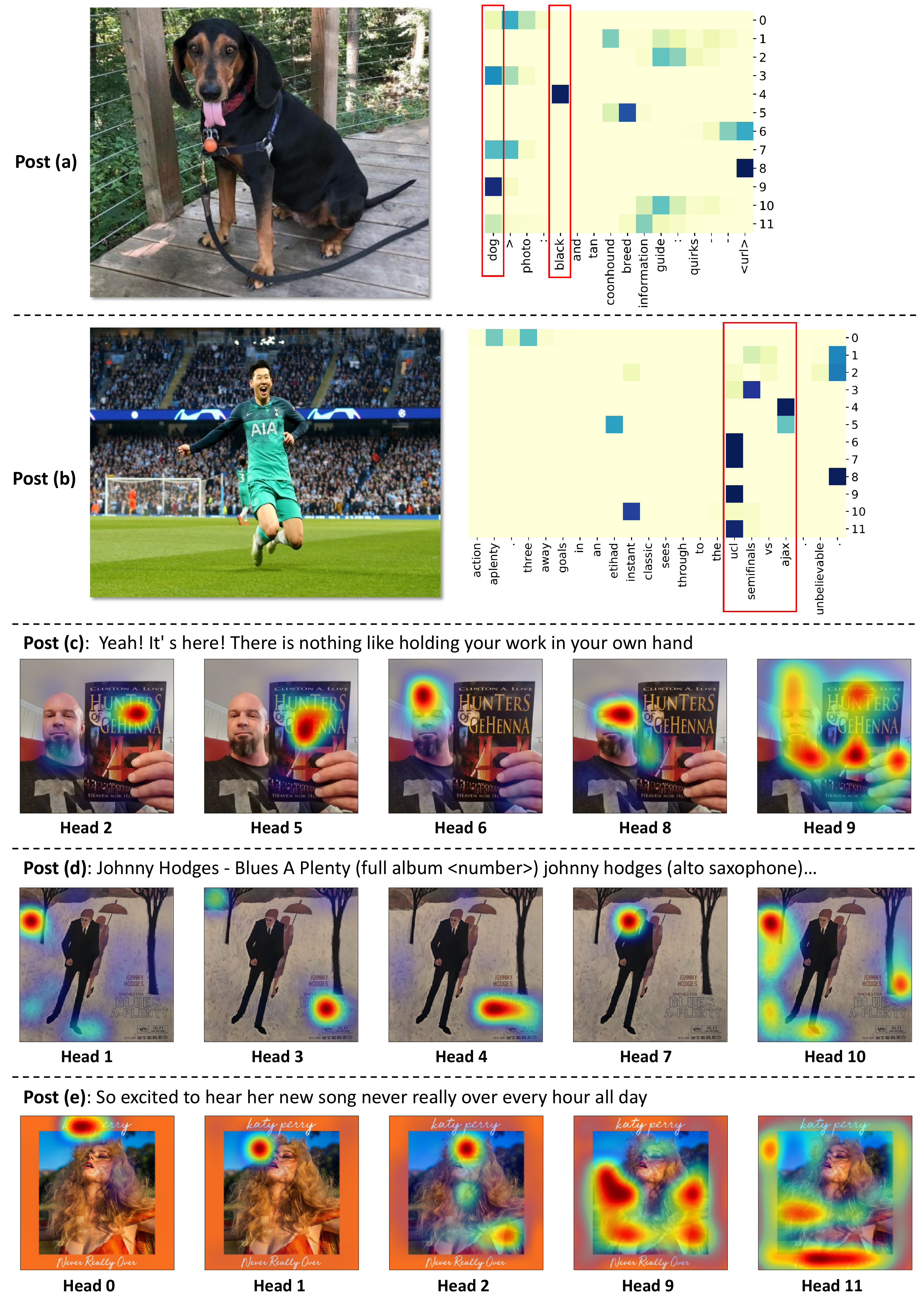}
\caption{More attention weight visualizations for both image-to-text attention and text-to-image attention.}\label{fig:more_attn_cases}
\end{figure*}

\begin{table*} [t]
\begin{center}
\resizebox{0.99\textwidth}{!}{

\begin{tabular}{p{4.8cm}p{4.8cm}p{4.8cm}p{4.8cm}}
\textbf{Post (a)}: Sharing is caring. Good girl Kit, cause I know how much you love your bed. \textit{\#Dogs} \textit{\#Kindness}
 &
\textbf{Post (b)}: Waves crash against the North Pier this evening at Tynemouth, River Tyne in the UK ~ @david1hirst \textit{\#StormHour}
&

 \textbf{Post (c)}: ``I am declaring an emergency that only i can fix''
 
 \textit{\#BoycottTrumpPrimeTime}
& 
\textbf{Post (d)}: The whole of the uk when armadillo and danny say anything
\textit{\#LoveIsland}\\

\begin{minipage}{\textwidth}
      \includegraphics[width=4.8cm,height=4.2cm]{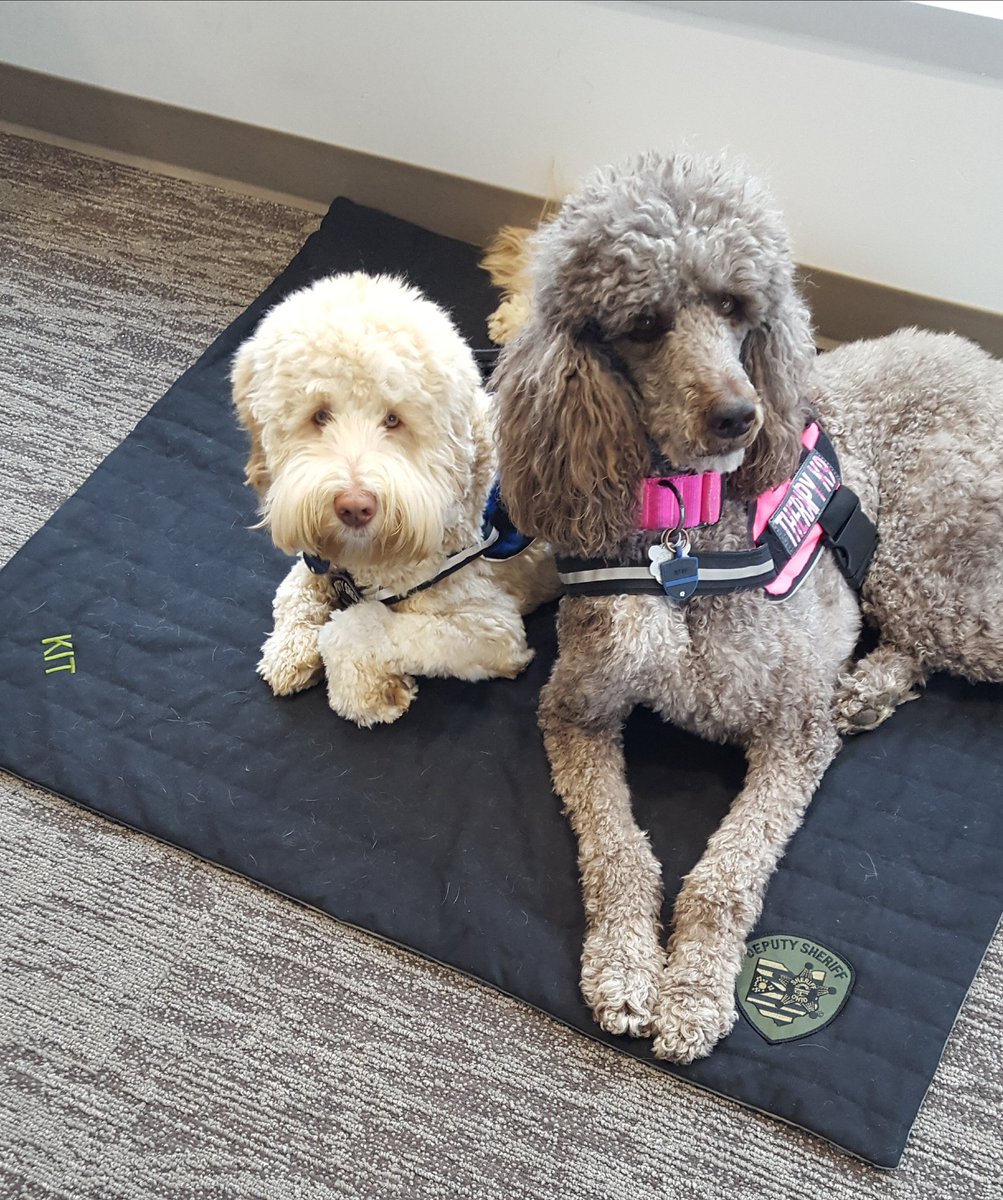}
\end{minipage}
 & 
 \begin{minipage}{\textwidth}
      \includegraphics[width=4.8cm,height=4.2cm]{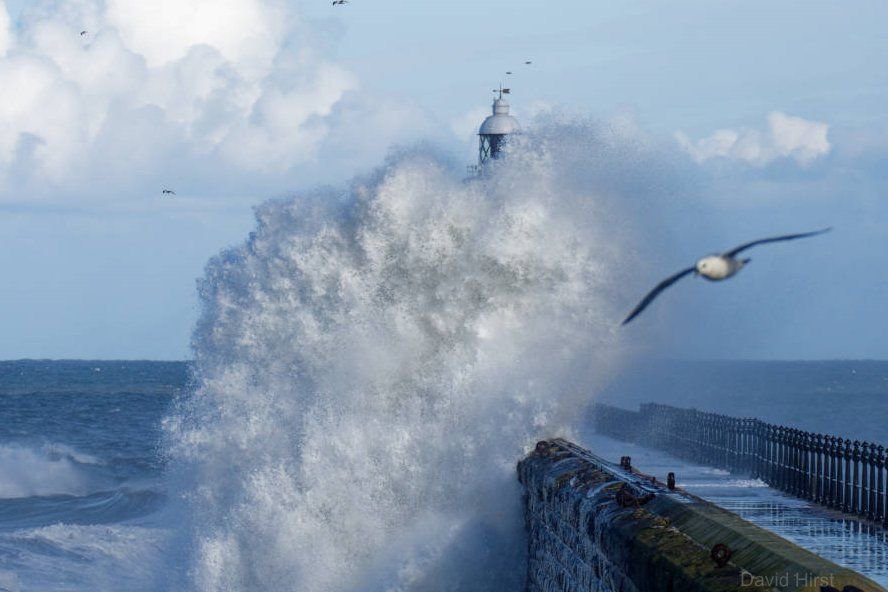}
\end{minipage}     
& 
 \begin{minipage}{\textwidth}
      \includegraphics[width=4.8cm,height=4.2cm]{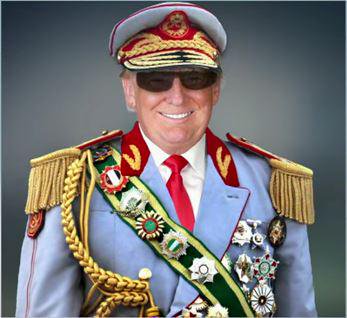}
 \end{minipage} 
& 
 \begin{minipage}{\textwidth}
      \includegraphics[width=4.8cm,height=4.2cm]{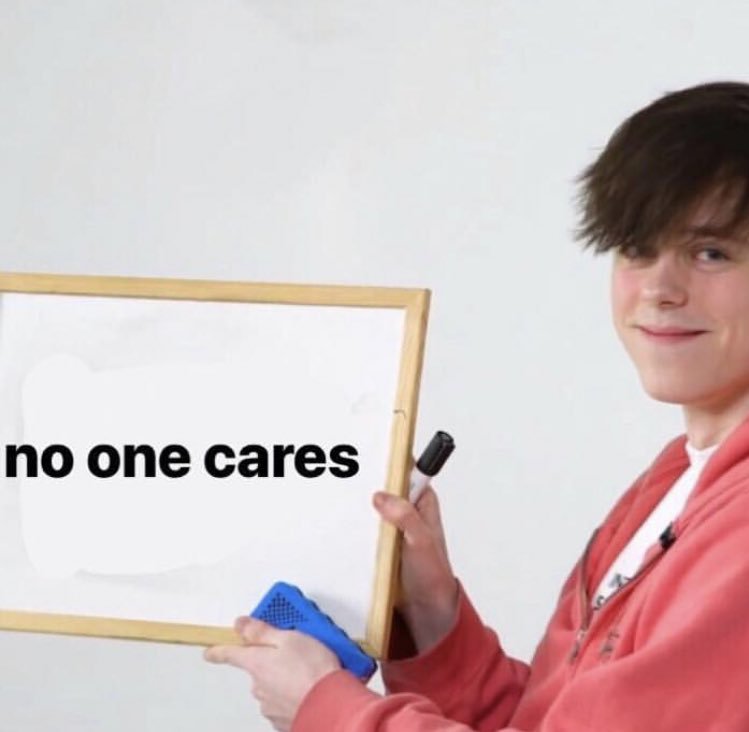}      
\end{minipage} \\

\end{tabular}
}
\end{center}
\captionsetup{type=figure}
\caption{Example tweets of four different types of text-image relationship in our dataset. Post (a): text is represented and image adds to.
Post (b): text is represented and image does not add to.
Post (c): text is not represented and image adds to.
Post (d): text is not represented and image does not add to.
}\label{tables:four_association_examples}

\end{table*}

\begin{table*} [t]
\begin{center}
\resizebox{0.99\textwidth}{!}{

\begin{tabular}{p{5.8cm}p{5.8cm}p{5.8cm}p{5.8cm}}
\textbf{Post (a)}: I thought Older Hanzo died after D'Vorah killed him? 
@NetherRealm~~~\textit{\#MortalKombat11}
 &
\textbf{Post (b)}: Congrats producer of the year, non-classical winner - Williams~~~\textit{\#Grammys}
&
\textbf{Post (c)}: Last year's highest rated animated movie spider man into the Spider-Verse is now streaming on Netflix!~~~\textit{\#SpiderMan}
&
\textbf{Post (d)}: We need to make sure the ratings are high

\textit{\#SaveShadowhunters}
\\
\begin{minipage}{\textwidth}
      \includegraphics[width=5.8cm,height=4cm]{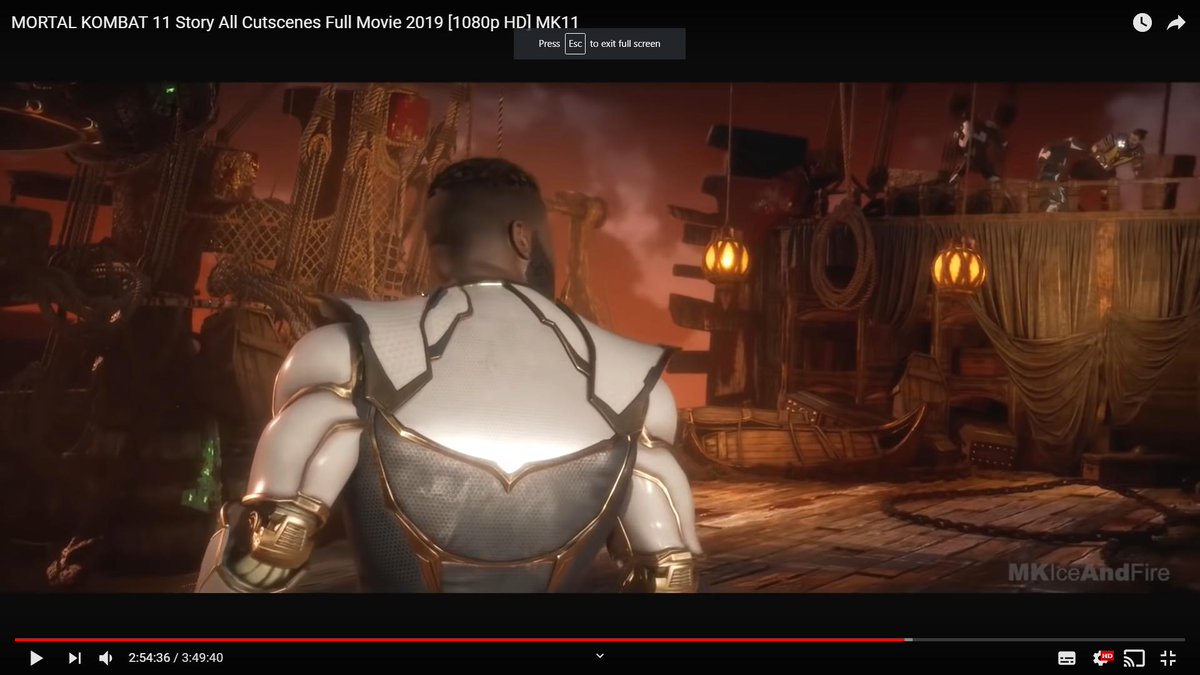}
\end{minipage}
 & 
 \begin{minipage}{\textwidth}
      \includegraphics[width=5.8cm,height=4cm]{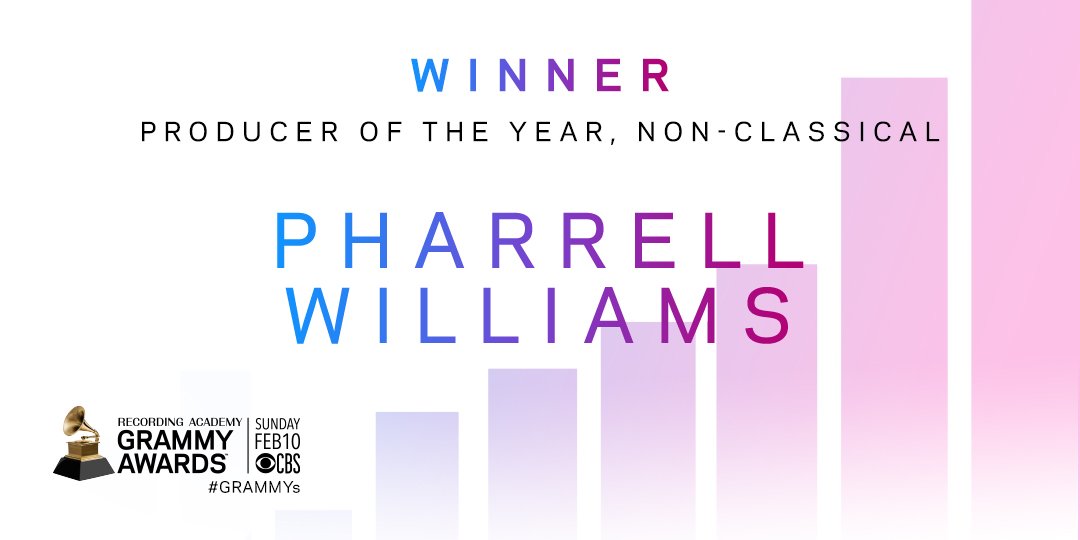}
\end{minipage}    
& 
 \begin{minipage}{\textwidth}
      \includegraphics[width=5.8cm,height=4cm]{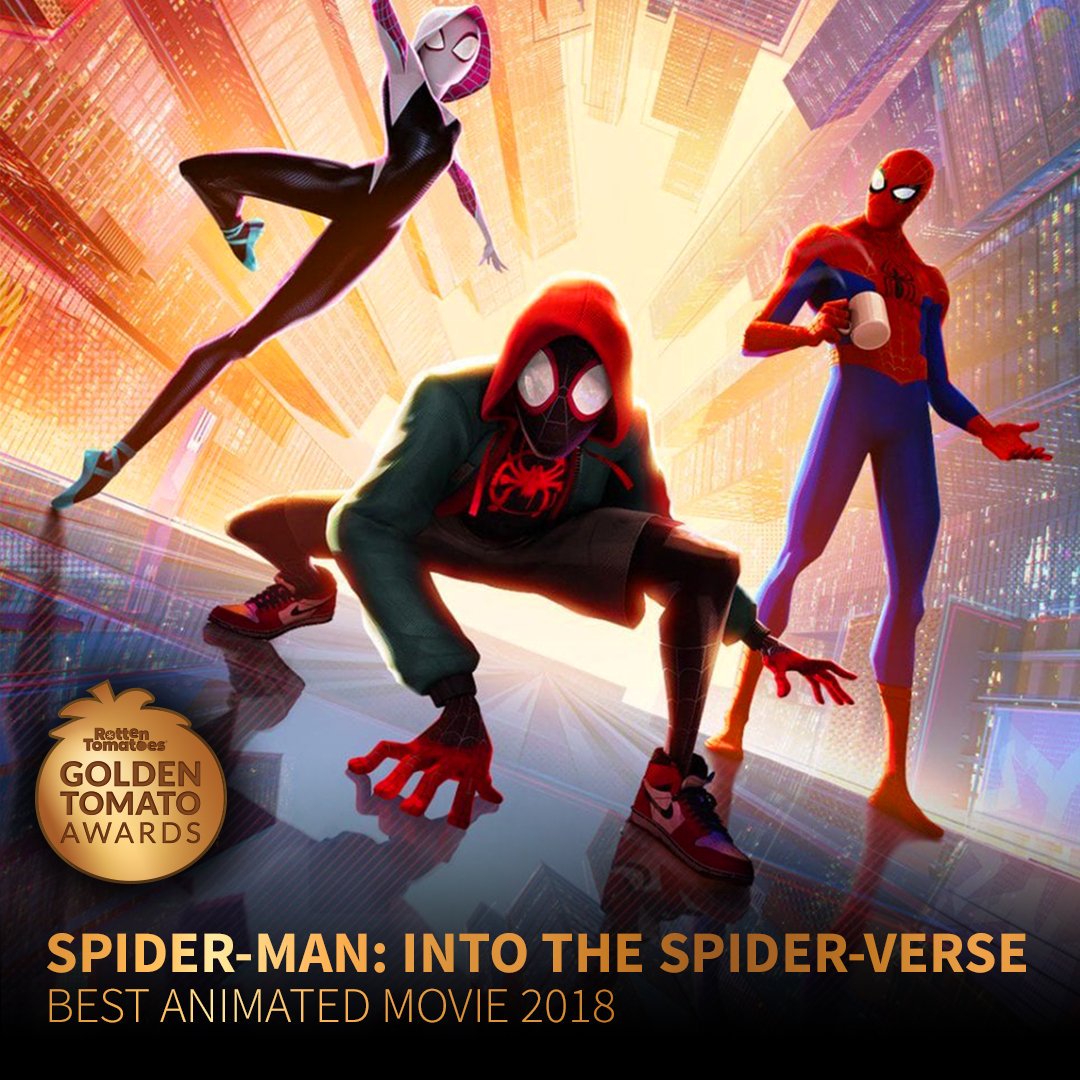}
 \end{minipage} 
& 
 \begin{minipage}{\textwidth}
      \includegraphics[width=5.8cm,height=4cm]{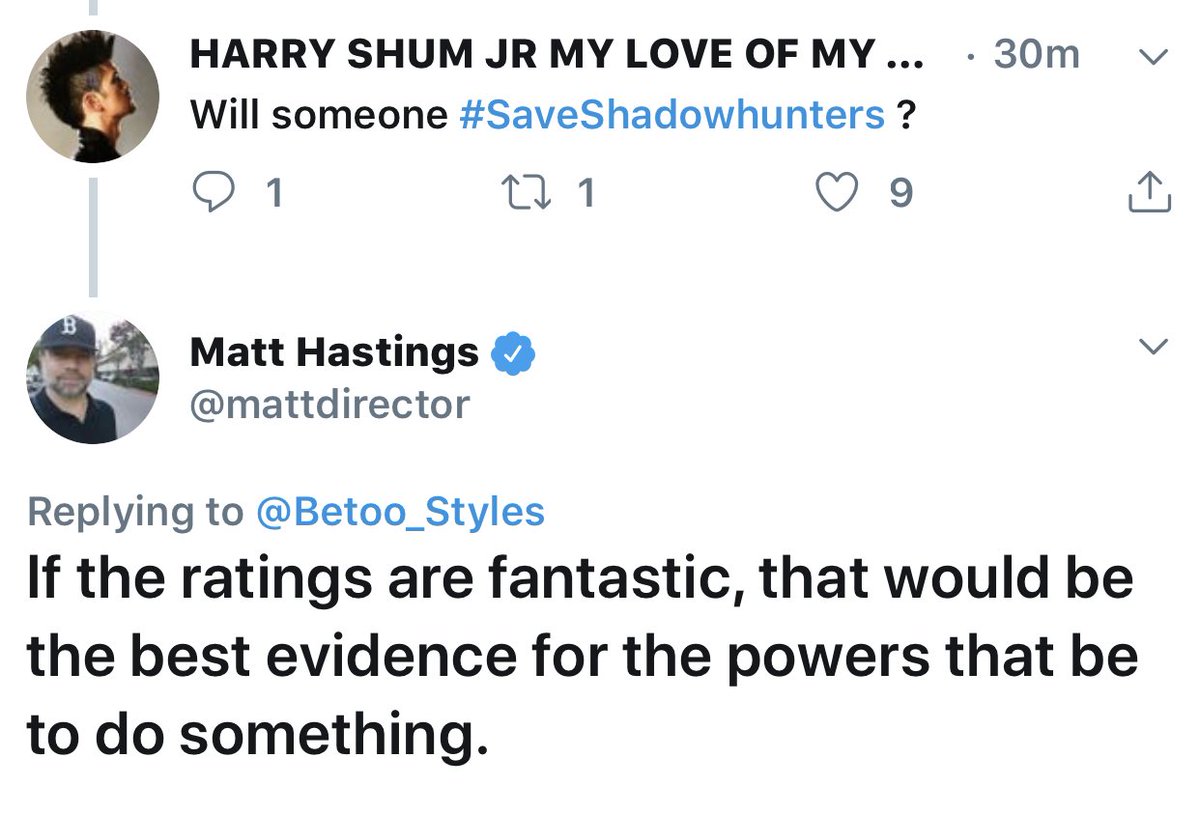} 
\end{minipage} 
\\
\multicolumn{1}{c}{(\textcolor{purple}{mortal kombat story all full movie})}
&
\multicolumn{1}{c}{(\textcolor{purple}{williams at grammy awards})}
&
\multicolumn{1}{c}{(\textcolor{purple}{spider man into the spider-verse})}
&
\multicolumn{1}{c}{(\textcolor{purple}{will someone save  shadow hunters})}
\\
\textsc{gen-copy}: quote

\textsc{cls-co-att}: destiny 2

Our: \textbf{mortal kombat 11}
&
\textsc{gen-copy}: live under par

\textsc{cls-co-att}: a star is born

Our: \textbf{grammys}
&
\textsc{gen-copy}: spider verse

\textsc{cls-co-att}: marvel

Our: \textbf{spider man}
&
\textsc{gen-copy}: teacher goals

\textsc{cls-co-att}: brexit

Our: \textbf{save shadowhunters}


\\

\end{tabular}

}
\end{center}
\captionsetup{type=figure}
\caption{More qualitative examples showing the effectiveness of encoding OCR texts. Among various models, only our model that considers OCR tokens  correctly predicts the keyphrases for all these cases (in bold). \textcolor{purple}{Purple tokens} are the detected OCR tokens, where we observe that the keyphrases directly appear in them.}\label{tables:OCR_cases}

\end{table*}

\begin{table*} [t]
\begin{center}
\resizebox{0.99\textwidth}{!}{

\begin{tabular}{p{5.2cm}p{5.2cm}p{5.2cm}p{5.2cm}}
\textbf{Post (a)}: Good night, everyone. I hope that you have had a delightful day and a restful weekend.~~\textit{\#hoorayfordogs} 
 &
\textbf{Post (b)}: Head up, chest out!  A handsome purple finch poses for a shot. 

\textit{\#birds} \textit{\#wildlife} \textit{\#photography}
&
\textbf{Post (c)}: I was watching all the bees Honeybee collecting pollen on the flowers Bouquet
\textit{\#SaveTheBees} \textit{\#CatsOfTwitter} 
&
\textbf{Post (d)}: For 1970, Plymouth intended to make its GTX model a street powerhouse.~~~\textit{\#MuscleCar} \textit{\#ClassicCar}
\\
\begin{minipage}{\textwidth}
      \includegraphics[width=5.2cm,height=4.5cm]{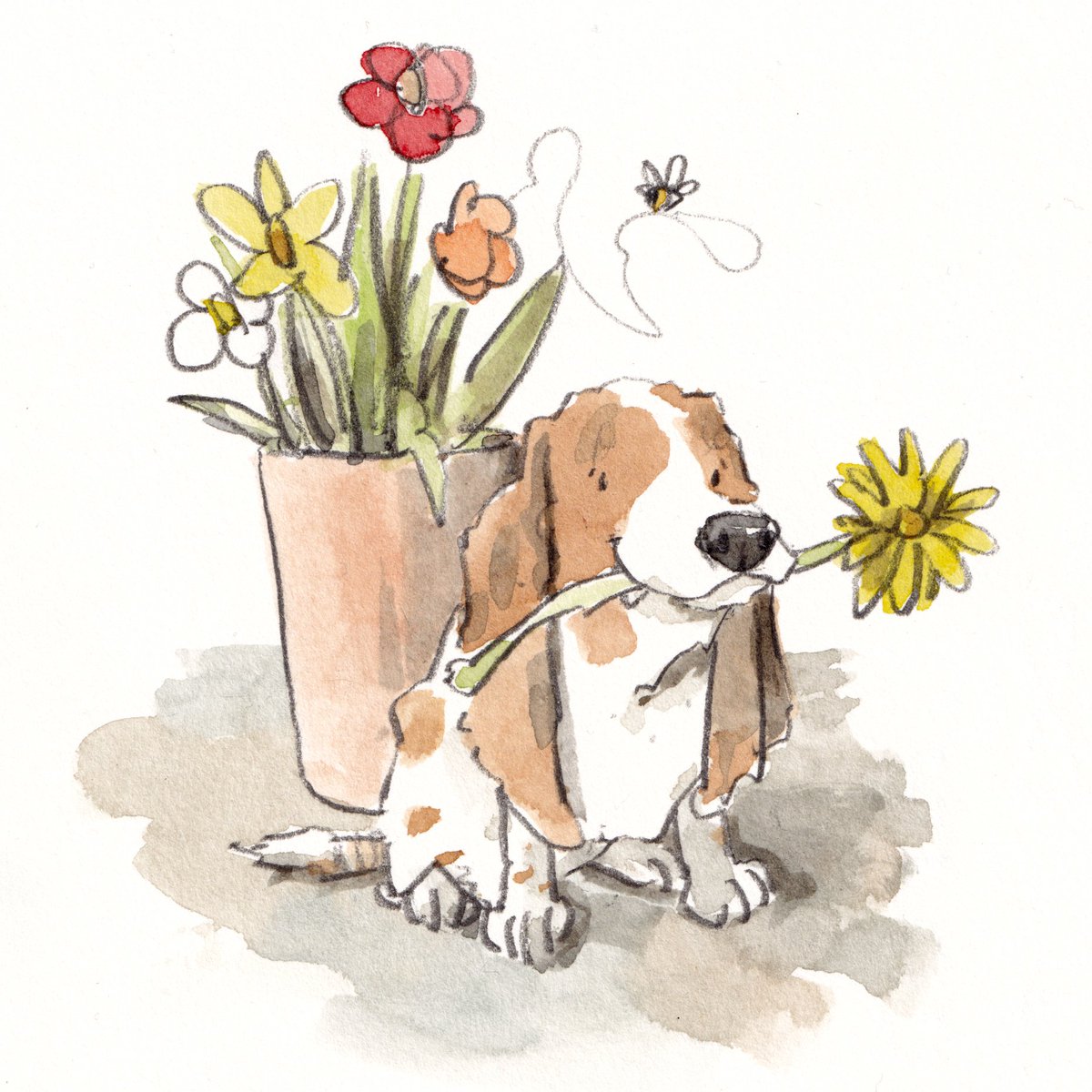}
\end{minipage}
 & 
 \begin{minipage}{\textwidth}
      \includegraphics[width=5.2cm,height=4.5cm]{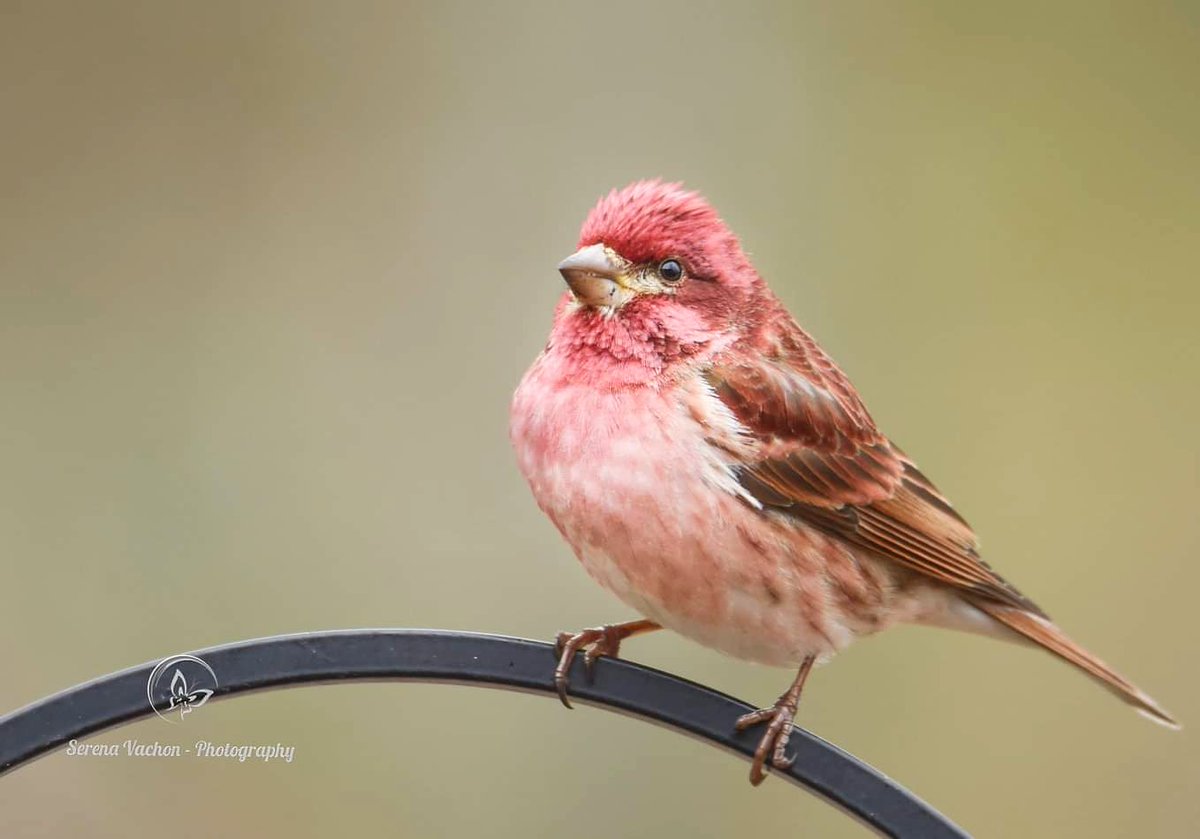}
\end{minipage}    
& 
\begin{minipage}{\textwidth}
      \includegraphics[width=5.2cm,height=4.5cm]{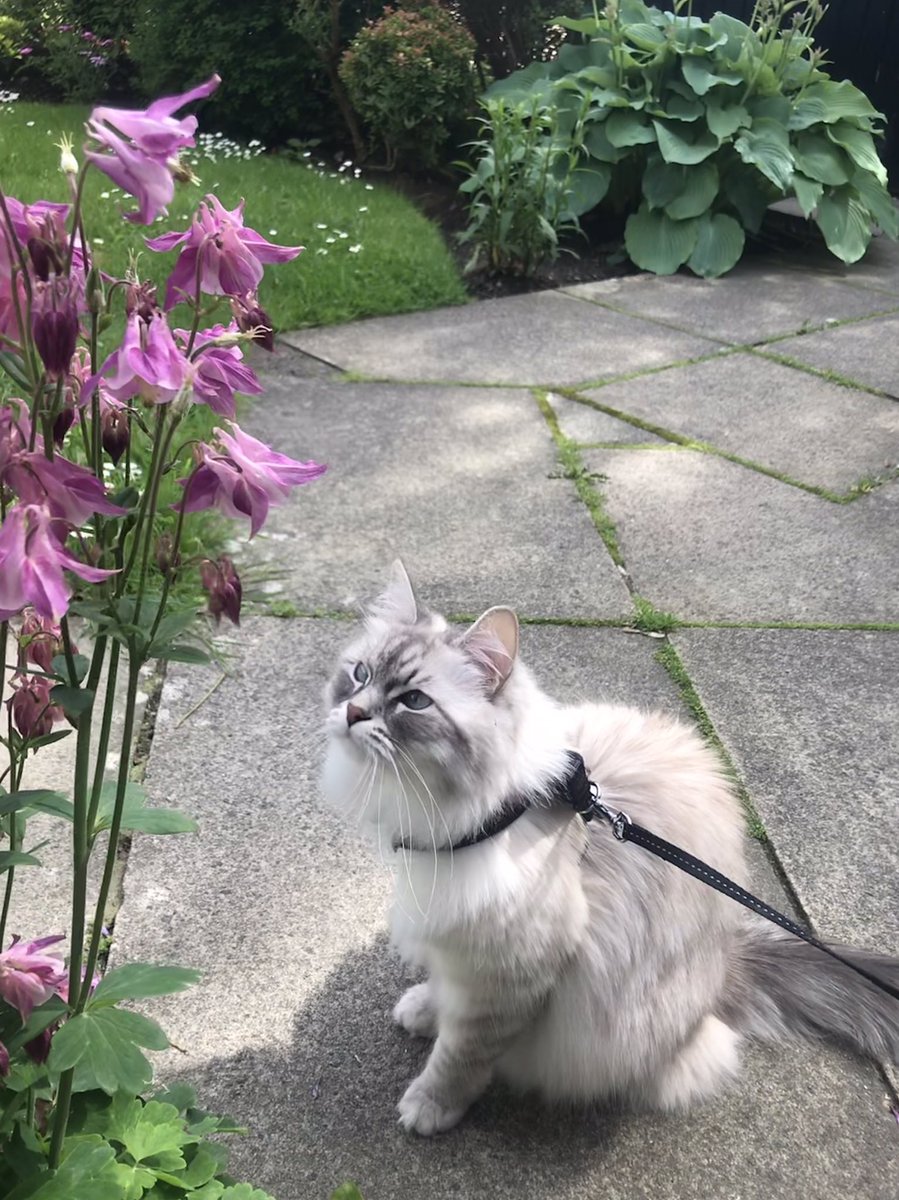}
\end{minipage} 
& 
 \begin{minipage}{\textwidth}
      \includegraphics[width=5.2cm,height=4.5cm]{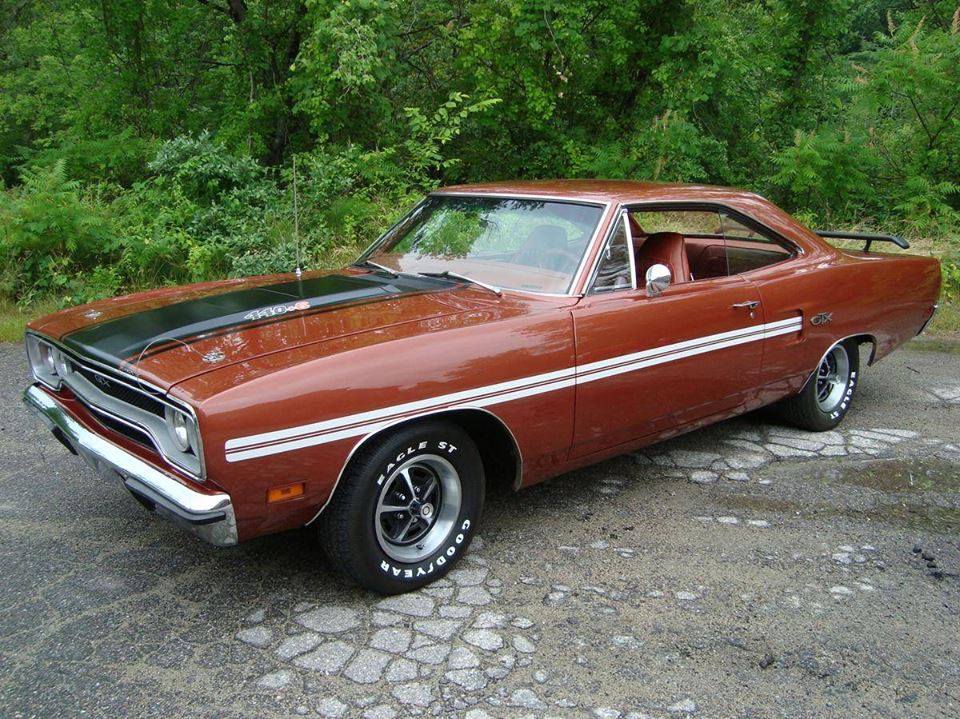}      
\end{minipage} 
\\
\multicolumn{1}{c}{(\textcolor{blue}{dog white yellow brown plate})}
&
\multicolumn{1}{c}{(\textcolor{blue}{branch bird red top small})}
&
\multicolumn{1}{c}{(\textcolor{blue}{cat white pink grey flowers})}
&
\multicolumn{1}{c}{(\textcolor{blue}{car roof park old meter})}
\\
\textsc{gen-copy}: friday feeling 

\textsc{cls-co-att}: \textbf{hooray for dogs}

Our: \textbf{hooray for dogs}
&
\textsc{gen-copy}: gap ol

\textsc{cls-co-att}: \textbf{birding}

Our: \textbf{birds}; \textbf{wildlife}
&
\textsc{gen-copy}: photography

\textsc{cls-co-att}: springwatch

Our: \textbf{cats of twitter}
&
\textsc{gen-copy}: plymouth

\textsc{cls-co-att}: mopar

Our: \textbf{classic car}
\\

\end{tabular}
}
\end{center}
\captionsetup{type=figure}
\caption{More qualitative examples showing the effectiveness of encoding image attributes. Our model that considers image attributes  correctly predicts the keyphrases for all these cases (in bold). \textcolor{blue}{Blue tokens} are the top five predicted attributes, which reveal the main image contents and thus help to indicate keyphrases.}\label{tables:attr_cases}

\end{table*}

\end{document}